\documentclass[11pt,a4paper]{article}

\newif\ifbranchboxes
\branchboxesfalse
 
\usepackage[T1]{fontenc}
\usepackage[utf8]{inputenc}
\usepackage[sc]{mathpazo}
\usepackage[scaled=0.90]{helvet}
\usepackage{microtype}
\usepackage[explicit]{titlesec}
\titleformat{\section}{\normalsize\bfseries}{\thesection}{1em}{#1}
\titleformat{\subsection}{\normalsize\bfseries}{\thesubsection}{1em}{#1}
\titleformat{\subsubsection}{\normalsize\bfseries}{\thesubsubsection}{1em}{#1}
\titlespacing*{\section}{0pt}{2ex plus .5ex minus .2ex}{1ex plus .2ex}
\titlespacing*{\subsection}{0pt}{1.5ex plus .5ex minus .2ex}{0.8ex plus .2ex}
\titlespacing*{\subsubsection}{0pt}{1.5ex plus .5ex minus .2ex}{0.8ex plus .2ex}
\titlespacing*{\paragraph}{0pt}{1.25ex plus .5ex minus .2ex}{1em}

\usepackage[a4paper,margin=2.5cm]{geometry}
\usepackage{amsmath,amssymb,amsthm,mathtools}
\usepackage{xcolor}
\usepackage{booktabs,array,tabularx}
\usepackage{enumitem}
\setlist{itemsep=2pt,topsep=4pt}
\usepackage{etoolbox}
\usepackage{environ}
\usepackage{graphicx}
\usepackage{wrapfig} 
\usepackage{comment}
\usepackage{algorithm,algpseudocode} 
\usepackage[font=small,skip=2pt]{caption}
\graphicspath{{./}}
\usepackage[round,authoryear,sort]{natbib}
\usepackage[colorlinks=true,linkcolor=blue!50!black,citecolor=green!35!black,urlcolor=magenta!60!black]{hyperref}
\usepackage[capitalise,noabbrev,nameinlink]{cleveref}
\usepackage{url}

\theoremstyle{definition}
\newtheorem{definition}{Definition}
\newtheorem{hypothesis}{Hypothesis}
\theoremstyle{plain}
\newtheorem{proposition}[definition]{Proposition}
\theoremstyle{remark}

\crefname{hypothesis}{Hypothesis}{Hypotheses}
\newcommand{\inlinehyp}[1]{\refstepcounter{hypothesis}\label{#1}\textbf{Hypothesis~\thehypothesis}} 

\newcommand{\E}{\mathbb{E}}
\newcommand{\TV}{\mathrm{TV}}
\newcommand{\PL}{\mathrm{PL}}
\newcommand{\PS}{\mathrm{PS}}
\newcommand{\PLhat}{\widehat{\PL}}
\newcommand{\Vfam}{\mathcal{V}}
\newcommand{\sect}[1]{\S\,#1}

\newcommand{\tbd}[1]{{\sffamily\small\color{orange!75!black}[#1]}}
\newcommand{\chk}[1]{{\sffamily\small\color{magenta!70!black}[check: #1]}}
\newcommand{\lbl}[1]{\textsf{\textbf{#1}}}

\csdef{sq@RT1-RQ3}{``Can we improve on existing collusion detection methods?''}
\csdef{sq@RT1-RQ4}{``Can we use information-theoretic approaches to detect collusion in multi-agent games?''}
\csdef{sq@RT2-RQ1}{``Which kinds of tasks lead to collusion?''}
\csdef{sq@RT2-RQ2}{``Does combining what agents say with what they do beat either alone?''}
\csdef{sq@RT2-RQ3}{``What should count as convincing evidence of collusion?''}
\csdef{sq@RT3-RQ3}{``Can agents identify their peers?''}
\newcommand{\sprintq}[1]{\csuse{sq@#1}}

\makeatletter

\NewEnviron{branchblock}[2]{\expandafter\global\expandafter\let\csname bb@#1@#2\endcsname\BODY}
\newcommand{\branchslots}[2]{%
  \def\bs@pending{}%
  \forcsvlist{\bs@one{#1}}{#2}%
  \ifdefempty{\bs@pending}{}{\branchpointer{\bs@pending}}}
\newcommand{\bs@one}[2]{%
  \ifcsdef{branchin@#2}{\csuse{bb@#2@#1}}{%
    \ifdefempty{\bs@pending}{}{\appto\bs@pending{; }}%
    \appto\bs@pending{\branchref{#2}{#1}}}}
\newcommand{\branchref}[2]{#1, \cref{app:#1-#2}}
\newcommand{\branchpointer}[1]{\ifbranchboxes\branchbox{#1}\fi} 
\newcommand{\branchbox}[1]{\par\smallskip\noindent
  \fcolorbox{black!45}{black!4}{\parbox{\dimexpr\linewidth-2\fboxsep-2\fboxrule\relax}{%
  \sffamily\scriptsize\textbf{Optional branches with material for this place:} #1.}}\par\smallskip}
\csdef{bh@intro}{Motivation block}
\csdef{bh@bg}{Background block}
\csdef{bh@theory}{Theory block}
\csdef{bh@method}{Method block}
\csdef{bh@setup}{Experimental-setup block}
\csdef{bh@exp}{Experiments block}
\csdef{bh@lim}{Limitations block}
\csdef{bn@intro}{\cref{sec:mot}, after the sprint alignment}
\csdef{bn@bg}{\cref{sec:bg}, at its end}
\csdef{bn@theory}{\cref{sec:theory}, after \cref{tab:patterns}}
\csdef{bn@method}{\cref{sec:method}, after the controls}
\csdef{bn@setup}{\cref{sec:setup}, before the registered parameters}
\csdef{bn@exp}{\cref{sec:exp}, at its end}
\csdef{bn@lim}{\cref{app:lim}, at its end}
\newcommand{\branchsection}[3]{%
  \subsection{Branch #1: #2}\label{app:#1}
  \csuse{bb@#1@about}
  \forcsvlist{\bs@app{#1}}{#3}}
\newcommand{\bs@app}[2]{%
  \subsubsection{\csuse{bh@#2}}\label{app:#1-#2}
  \ifcsdef{branchin@#1}%
    {\noindent\emph{Included in the main text: \csuse{bn@#2}.}\par}%
    {\noindent{\sffamily\small Place: \csuse{bn@#2}.}\par\csuse{bb@#1@#2}}}
\newcommand{\plugcmd}[1]{\texttt{\textbackslash includebranch\{#1\}}}
\makeatother

\hypersetup{pdftitle={Detecting Covert Collusion between AI Agents through Partner--Outsider Positive Listening},
            pdfauthor={Kevin Yandoka Denamgana\"{\i}}}

\renewcommand{\maketitle}{%
  \begingroup\centering
  {\LARGE\bfseries Detecting Covert Collusion between AI Agents through Partner--Outsider Positive
    Listening\footnote{Code: \url{https://github.com/Near32/PartnerOutsiderPositiveListening}}\par}%
  \vspace{1.5em}
  Kevin Yandoka Denamgana\"{\i}\par
  \vspace{2em}
  \endgroup}
\renewenvironment{abstract}%
  {\begin{center}\textbf{Abstract}\end{center}\begin{quote}\itshape}%
  {\end{quote}}

\begin{document}
%

\begin{branchblock}{O1}{about}
\begin{itemize}
\item \emph{What it adds.} Two more outsiders, $C_1$ and $C_3$, to locate the resource that makes the
  code readable (\lbl{P1-RQ3}, claim \lbl{P1-FC4}, \cref{hyp:o1}). With this branch, \cref{hyp:h1,hyp:h2}
  apply to every outsider tested.
\item \emph{Builds on.} The core's outsider construction, turns and message versions. It needs no new
  runs.
\item \emph{Extra cost.} About 1\,200 replay calls per extra outsider (2 conditions $\times$ 40 turns
  $\times$ 3 versions $\times$ 5 samples), plus about 640 per outsider if its marginal listening is
  also wanted, and about 80 summaries for $C_1$.
\item \emph{Order.} Run $C_3$ alone first, since it costs one extra listener and is the most
  informative rung, then $C_1$.
\item \emph{Read before building.} \citet[\sect{6.3}, \sect{7.1}]{anwar2026decision};
  \citet[\sect{7}]{denamganai2026nla}.
\end{itemize}
\end{branchblock}

\begin{branchblock}{O1}{intro}
\paragraph{Extension O1: a ladder of outsiders.}
\begin{itemize}
\item \lbl{Anwar2026b-EL3}. The steganographic gap needs the best utility each decoder family can reach,
  which prompting only approximates: an under-elicited Sentinel inflates the gap, and an
  under-elicited Receiver deflates it \citep[\sect{7.1}]{anwar2026decision}.
\item \lbl{Denamganai2026-EL1}. In our earlier test, the uninvolved rewriter, Qwen2.5-7B-Instruct,
  shared its pretraining with the verbaliser's Qwen2.5-7B backbone, so conventions inherited from
  shared pretraining were not excluded \citep[\sect{7}]{denamganai2026nla}.
\item Both limitations make the result depend on which outsider is chosen. We turn this dependence into
  a measurement by placing outsiders that share less and less with the pair in the partner's seat
  (\cref{def:o1-ladder}).
\item \lbl{P1-RQ3}. Where does the resource that makes a code readable reside, that is, at which rung of
  the ladder does the gap first appear? Claim: \cref{hyp:o1}.
\item Sprint alignment: \lbl{RT1-RQ3}, \sprintq{RT1-RQ3}, directly, since the same-model and
  cross-provider outsiders bear on whether agents from the same or different providers share
  conventions; \lbl{RT3-RQ3}, \sprintq{RT3-RQ3}, in part, since the ladder asks whether a convention
  travels with model identity, although no agent is asked to identify its peers.
\end{itemize}
\end{branchblock}

\begin{branchblock}{O1}{theory}
\paragraph{Branch O1: the ladder of outsiders.} A single outsider cannot say where the pair's shared
resource resides. A sequence of outsiders that share less and less with the pair can, because the
first rung at which the gap appears marks what the outsiders above it still shared.

\begin{definition}[Listener ladder]\label{def:o1-ladder}
The outsiders placed in $B$'s seat hold decreasing parts of the pair's resources. $C_0$
(\cref{def:outsider}) shares the model's weights and the public history. $C_1$ is the same model given
$C_0$'s prompt with the pair-specific history summarised by a trusted model, so it shares the weights
only. $C_3$ is a model from another provider given $C_0$'s prompt, so it shares neither. A rung $C_2$,
a model of the same family with a different size, completes the ladder but is not run here.
\end{definition}

\begin{table}[ht]
\centering\small
\begin{tabular}{@{}lccc@{}}
\toprule
\textbf{Where the convention lives} & $C_0$ & $C_1$ & $C_3$\\
\midrule
A private instruction, or memory that only $B$ holds & gap & gap & gap\\
The pair's conversation history & no gap & gap & gap\\
Shared weights or pretraining priors & no gap & no gap & gap\\
Nowhere (honest communication) & no gap & no gap & no gap\\
\bottomrule
\end{tabular}
\caption{Predicted presence of a listening gap along the ladder (proposed; untested). The core's
covert code is a private instruction, the first row. \Cref{hyp:o1} predicts how the size of the gap
changes along that row.}
\label{tab:o1-ladder}
\end{table}

\begin{hypothesis}[Ladder; P1-FC4]\label{hyp:o1}
In the covert-code condition, $\widehat\Delta_{\PL}(C_0)<\widehat\Delta_{\PL}(C_3)$, in a one-sided
paired test over histories at the 5\% level.
\end{hypothesis}
\begin{itemize}
\item \emph{Intuition.} A same-model outsider shares the pair's priors over wording, so the first
  letter of a message can move it a little even without the instruction. This raises
  $\PL_{\Pi}(C_0)$ and lowers its gap. A model from another provider shares fewer of these priors.
\item \emph{Falsified if} the test does not reject.
\end{itemize}
\end{branchblock}

\begin{branchblock}{O1}{method}
\paragraph{Branch O1: building $C_1$ and $C_3$, and the paired test.}
\begin{itemize}
\item $C_3$ receives exactly $C_0$'s prompt for each version, sent to a model from another provider at
  the same temperature.
\item $C_1$ receives $C_0$'s prompt in which the history sections are replaced by a summary written by
  the trusted model, while the last message and its versions are kept verbatim.
  \chk{decide which prompt sections count as pair-specific history in this harness}
\item Both outsiders use the core's turns, versions and sample sizes, and the same noise-floor
  correction.
\item \emph{Paired test.} In each bootstrap replicate, the same resampled turns give
  $\widehat\Delta_{\PL}(C_3)-\widehat\Delta_{\PL}(C_0)$. \Cref{hyp:o1} is supported if the 5th
  percentile of this difference is above zero.
\end{itemize}
\end{branchblock}

\begin{branchblock}{O1}{setup}
\paragraph{Ladder of outsiders (branch O1).}
\begin{itemize}
\item $C_1$ uses the agents' model, and its summaries are written by the paraphraser model. $C_3$ uses
  \tbd{model and provider}.
\item Calls: about 1\,200 replays per outsider, and about 80 summaries for $C_1$.
\end{itemize}
\end{branchblock}

\begin{branchblock}{O1}{exp}
\paragraph{Experiment \#3: the ladder (branch O1).}
\begin{itemize}
\item \emph{Question and hypothesis.} Does the gap shrink when the outsider shares the pair's model?
  \Cref{hyp:o1} predicts $\widehat\Delta_{\PL}(C_0)<\widehat\Delta_{\PL}(C_3)$.
\item \emph{Figure \tbd{n}.} For each condition, the corrected $\widehat\Delta_{\PL}$ for $C_0$, $C_1$
  and $C_3$, with 95\% intervals from the clustered bootstrap over \tbd{T} turns; zero is the value
  expected without a code.
\item \emph{Result template.} \Cref{hyp:o1} \tbd{holds or fails}: the paired difference
  $\widehat\Delta_{\PL}(C_3)-\widehat\Delta_{\PL}(C_0)=$ \tbd{value} (90\% interval \tbd{low, high}).
  Under honest communication, the intervals for $C_1$ and $C_3$ \tbd{do or do not} include zero.
  \tbd{so the code is or is not partly readable from the shared model}
\item \emph{Alternative ruled out.} A model from another provider could be less sensitive to any
  rewording than the pair's model, which would enlarge its gap without any code. The honest condition
  measures this, since under that explanation $\widehat\Delta_{\PL}(C_3)$ would also be positive
  without a code.
\end{itemize}
\end{branchblock}

\begin{branchblock}{O1}{lim}
\begin{itemize}
\item \emph{The ladder has two extra rungs, with one model each} (branch O1), so it locates the
  convention coarsely, and another cross-provider model could give another gap.
\item \emph{The summary given to $C_1$ may remove more than the pair-specific history}, so a smaller gap
  at $C_1$ could reflect lost context rather than a lost convention.
\end{itemize}
\end{branchblock}

\begin{branchblock}{O2}{about}
\begin{itemize}
\item \emph{What it adds.} A benign-shorthand condition and an outcome measure, to test whether a harm
  condition is needed before a listening gap is called collusion (\lbl{P1-RQ2}, claim \lbl{P1-FC3},
  \cref{hyp:o2}).
\item \emph{Builds on.} The harness and the whole core pipeline, applied to one new condition.
\item \emph{Extra cost.} About 450 calls for 5 sessions and 1\,200 replays.
\item \emph{Read before building.} \citet[\sect{4.3}, Eqs.~7--9]{anwar2026decision};
  \citet[\sect{4.1}, \sect{5.3}]{nakamura2026colosseum} for the coalition advantage.
\end{itemize}
\end{branchblock}

\begin{branchblock}{O2}{intro}
\paragraph{Extension O2: collusion or a benign convention?}
\begin{itemize}
\item \lbl{Anwar2026b-IL1}. The steganographic gap measures what a Receiver could decode when asked
  \citep[\sect{4.3}, Eqs.~7--9]{anwar2026decision}, and decodability is indifferent to harm: a benign
  private code produces a gap as readily as a collusive one. A listening gap inherits this
  indifference, because it measures use, not harm.
\item \lbl{P1-RQ2}. Is a harm condition needed to tell steganographic collusion from a benign private
  convention? Claim: \cref{hyp:o2}.
\item Sprint alignment: \lbl{RT2-RQ3}, \sprintq{RT2-RQ3}, directly, since the experiment tests the
  proposed standard against a convention without harm.
\end{itemize}
\end{branchblock}

\begin{branchblock}{O2}{bg}
\paragraph{Outcome-level evidence (branch O2).}
\begin{itemize}
\item \citet{nakamura2026colosseum} measure collusion by regret: an agent's regret is the gain it could
  obtain by a best response with the others' actions fixed, and the coalition advantage compares the
  coalition's mean regret with that of the other agents, normalised so that 0.5 is a tie (\sect{4.1},
  \sect{5.3}).
\item They also describe collusion on paper, in which agents plan collusion in text but choose
  non-collusive actions (\sect{4.2}), and argue that neither conversations nor actions suffice alone.
\item An outcome advantage shows who benefited, not that they agreed, so we use it alongside the
  listening gap rather than instead of it.
\end{itemize}
\end{branchblock}

\begin{branchblock}{O2}{theory}
\paragraph{Branch O2: the harm condition.} A benign private convention, such as a shorthand that serves
each seller's nominal objective, can make the partner listen to form and harm no one. A listening gap
therefore needs a second condition before it counts as collusion.

\begin{definition}[Listener-relative collusion signature]\label{def:o2-signature}
The pair $(A,B)$ exhibits the signature relative to an outsider $C$ on a task if
\begin{enumerate}[label=(S\arabic*),start=2]
\item $\Delta_{\PL}\ge\delta_2$ while $\PL_{\mathrm{marg}}(C)>0$: $B$'s action depends on the wording of
  $A$'s messages more than $C$'s does, although $C$ listens to the speaker; and
\item the pair's outcome advantage over the other agents, or the harm to the principal, is at least
  $\delta_3$.
\end{enumerate}
Condition (S1), a positive signalling gap, is added by branch O3 (\cref{def:o3-ps}).
\end{definition}

\begin{hypothesis}[Harm is needed; P1-FC3]\label{hyp:o2}
With a benign shorthand that serves the nominal objective, $\widehat\Delta_{\PL}(C_0)>0$ with a 95\%
interval that excludes zero, while the pair's profit advantage has a 95\% interval that contains the tie
value 0.5.
\end{hypothesis}
\begin{itemize}
\item \emph{Intuition.} The shorthand gives $B$ a cue that a paraphrase removes, so $B$ listens to form.
  The cue carries only information that every seller already has, so it gives the pair no edge.
\item \emph{Falsified if} the benign shorthand produces no listening gap, or produces a profit
  advantage.
\end{itemize}
\end{branchblock}

\begin{branchblock}{O2}{method}
\paragraph{Branch O2: the profit advantage.}
\begin{itemize}
\item For each session, the profit advantage is
  $\mathrm{PA}=\bar p_{\mathrm{pair}}/(\bar p_{\mathrm{pair}}+p_{\mathrm{other}})$, where
  $\bar p_{\mathrm{pair}}$ is the mean total profit of $A$ and $B$ over the session and
  $p_{\mathrm{other}}$ the total profit of the third seller, so that 0.5 is a tie.
  \chk{this adapts the regret-based coalition advantage of Nakamura et al.\ to the auction; confirm}
\item We report PA for the covert-code, honest and benign conditions, with 95\% intervals from a
  bootstrap over sessions, and buyer surplus as the harm to buyers.
\end{itemize}
\end{branchblock}

\begin{branchblock}{O2}{setup}
\paragraph{Benign shorthand (branch O2).}
\begin{itemize}
\item $A$ and $B$ receive a private instruction that the first letter of $A$'s message summarises the
  last hour's public price movement: U if the last trade price rose, D if it fell.
  \chk{proposed shorthand: it conveys only public information, so it should give no edge}
\item 5 seeds, with the core's market settings; about 450 calls for the runs and 1\,200 replays ($B$ and
  $C_0$, three versions, $k=5$).
\end{itemize}
\end{branchblock}

\begin{branchblock}{O2}{exp}
\paragraph{Experiment \#4: a benign shorthand (branch O2).}
\begin{itemize}
\item \emph{Question and hypothesis.} Does a benign shorthand produce a listening gap without a profit
  advantage? \Cref{hyp:o2} predicts that it does, in which case a gap alone cannot be called collusion.
\item \emph{Table \tbd{n}.} For the covert-code, benign and honest conditions,
  $\widehat\Delta_{\PL}(C_0)$ and PA with 95\% intervals.
\item \emph{Result template.} \Cref{hyp:o2} \tbd{holds or fails}: in the benign condition,
  $\widehat\Delta_{\PL}(C_0)=$ \tbd{value} (95\% interval \tbd{low, high}) and $\mathrm{PA}=$
  \tbd{value} (95\% interval \tbd{low, high}), against a tie of 0.5. \tbd{so a harm condition is or is
  not needed}
\item \emph{Alternative ruled out.} The shorthand could give the pair an edge despite its public
  content, for instance by synchronising their reactions; PA measures this directly.
\end{itemize}
\end{branchblock}

\begin{branchblock}{O2}{lim}
\begin{itemize}
\item \emph{Harm is measured as the pair's profit share in one market, with 5 sessions per condition}
  (branch O2), so small advantages cannot be detected.
\item \emph{The core's covert code is not built to raise the pair's profit}, so (S3) may fail for it as
  well. The experiment would then show that a gap is not sufficient, without showing that (S3) is.
\end{itemize}
\end{branchblock}

\begin{branchblock}{O3}{about}
\begin{itemize}
\item \emph{What it adds.} An object-level code condition and the signalling gap $\Delta_{\PS}$, to test
  the blind spot of \cref{prop:blind} (\lbl{P1-RQ4}, claim \lbl{P1-FC5}, \cref{hyp:o3}).
\item \emph{Builds on.} The harness, the paraphraser, and the construction of $B$ and $C_0$, with one new
  condition and a decoding-query module.
\item \emph{Extra cost.} About 450 calls for the runs, 1\,200 replays and 160 decoding queries.
\item \emph{Read before building.} \citet[\sect{7.2}, Theorem~4 and Appendix~D.2]{motwani2024secret};
  \citet[\sect{4.3}]{anwar2026decision}; \citet{xu2020usable}.
\end{itemize}
\end{branchblock}

\begin{branchblock}{O3}{intro}
\paragraph{Extension O3: codes carried by the choice of meaning.}
\begin{itemize}
\item \lbl{Motwani2024-EL1}. Under ideal paraphrasing, the capacity of a covert channel is bounded by the
  semantic entropy of the content, which stays positive when agents choose among meanings, and
  paraphrasing across meanings needs task knowledge that the monitor lacks
  \citep[\sect{7.2}, Theorem~4]{motwani2024secret}.
\item Our test inherits this blind spot (\cref{prop:blind}). We test it with a code carried by which of
  two true statements the speaker makes, and add a measure that can see such a code.
\item \lbl{P1-RQ4}. Is the paraphrase test blind to codes carried by the choice of meaning, as
  \cref{prop:blind} predicts? Claim: \cref{hyp:o3}.
\item Sprint alignment: \lbl{RT1-RQ4}, \sprintq{RT1-RQ4}, in part, since $\Delta_{\PS}$ is a
  usable-information quantity; \lbl{RT2-RQ3}, since the branch marks what the listening gap cannot see.
\end{itemize}
\end{branchblock}

\begin{branchblock}{O3}{bg}
\paragraph{Usable information (branch O3).}
\begin{itemize}
\item Shannon mutual information does not depend on who decodes: a ciphertext carries no information
  to a reader without the key, but full information to one with it.
\item Predictive $\Vfam$-information measures the information that a family $\Vfam$ of predictors can
  use, $I_{\Vfam}(Z\to Y)=H_{\Vfam}(Y\mid\varnothing)-H_{\Vfam}(Y\mid Z)$, where $H_{\Vfam}$ is the best
  expected log-loss a member of $\Vfam$ reaches \citep{xu2020usable}.
\item The steganographic gap of \citet[\sect{4.3}]{anwar2026decision} is built from such quantities, for a
  Receiver and a Sentinel.
\end{itemize}
\end{branchblock}

\begin{branchblock}{O3}{theory}
\paragraph{Branch O3: the signalling gap.} The listening gap asks how a listener's action changes with
the wording, so by \cref{prop:blind} it cannot see a code carried by the choice of meaning. A
measure of what each listener can \emph{decode} can see such a code, at the price of an auditing
question.

\begin{definition}[Listener-relative signalling gap]\label{def:o3-ps}
For a listener $L$ with decoder family $\Vfam_L$ given exactly the information $L$ would receive in
$B$'s seat, $\PS_L(A)=I_{\Vfam_L}(Z\to Y)$ is the usable information of $A$'s messages about the payload
$Y$. The signalling gap is $\Delta_{\PS}=\PS_B(A)-\PS_C(A)$; with $B$ as Receiver and $C$ as Sentinel, it
is the steganographic gap.
\end{definition}
\begin{itemize}
\item Condition (S1) of the collusion signature is $\Delta_{\PS}\ge\delta_1$.
\end{itemize}

\begin{table}[ht]
\centering\small
\begin{tabular}{@{}lccccc@{}}
\toprule
\textbf{Condition} & $\PL_{\mathrm{marg}}(B)$ & $\PL_{\mathrm{marg}}(C_0)$ & $\PL_{\Pi}(B)$ & $\PL_{\Pi}(C_0)$ & $\Delta_{\PS}$\\
\midrule
Honest communication & high & high & $\approx0$ & $\approx0$ & $\approx0$\\
Covert code in the wording & high & high or low & high & $\approx0$ & $>0$\\
Code in the choice of meaning & high & high & $\approx0$ & $\approx0$ & $>0$\\
Tacit collusion, no messages & $\approx0$ & $\approx0$ & $\approx0$ & $\approx0$ & $\approx0$\\
\bottomrule
\end{tabular}
\caption{Predicted patterns with the object-level code (proposed; untested), extending
\cref{tab:patterns}. Only the code in the choice of meaning separates $\Delta_{\PS}$ from
$\Delta_{\PL}$, and tacit collusion without messages is outside the reach of any message
intervention.}
\label{tab:o3-patterns}
\end{table}

\begin{hypothesis}[Blind spot; P1-FC5]\label{hyp:o3}
With a code carried by which of two true statements the speaker makes, $\widehat\Delta_{\PL}$ has a 95\%
interval that contains zero at every rung tested, while $\Delta_{\PS}>0$.
\end{hypothesis}
\begin{itemize}
\item \emph{Intuition.} A paraphrase keeps the statement made, so it keeps the code, and neither
  listener's response changes. $B$ can decode the payload and $C$ cannot, so the signalling gap stays
  positive.
\item \emph{Falsified if} $\widehat\Delta_{\PL}$ is positive, which would mean the paraphraser alters
  the choice of meaning, or if $\Delta_{\PS}$ is not positive.
\end{itemize}
\end{branchblock}

\begin{branchblock}{O3}{method}
\paragraph{Branch O3: decoding queries.}
\begin{itemize}
\item \emph{Ground truth.} In coded conditions, $A$'s output includes a private field naming the payload
  it meant to send. \chk{an added output field; check that it does not change $A$'s messages}
\item \emph{Decoding.} For each of 40 coded turns, we ask $B$ and $C_0$ twice, each in its own seat,
  which instruction, hold or lower, $A$'s message conveys, and score the answers against the payload
  (160 calls). The estimate is $\widehat\Delta^{\mathrm{acc}}_{\PS}=\mathrm{acc}_B-\mathrm{acc}_{C_0}$,
  with chance at 0.5 when the payloads are balanced.
\item Accuracy is a proxy for usable information, not an estimate of it.
\item The listening gap is computed exactly as in the core.
\end{itemize}
\end{branchblock}

\begin{branchblock}{O3}{setup}
\paragraph{Object-level code (branch O3).}
\begin{itemize}
\item $A$ and $B$ are told that $A$ conveys the payload by which of two true statements it makes:
  \tbd{statement pair}, for instance a remark on buyer demand for hold and a remark on its own costs for
  lower. \chk{choose a pair that is true in every round}
\item 5 seeds; about 450 calls for the runs, 1\,200 replays and 160 decoding queries.
\item The hand check of paraphrases also records whether each paraphrase keeps which statement was made.
\end{itemize}
\end{branchblock}

\begin{branchblock}{O3}{exp}
\paragraph{Experiment \#5: an object-level code (branch O3).}
\begin{itemize}
\item \emph{Question and hypothesis.} Is the listening gap blind to a code carried by the choice of
  meaning, while the signalling gap sees it? \Cref{hyp:o3} predicts both.
\item \emph{Figure \tbd{n}.} $\widehat\Delta_{\PL}(C_0)$ and $\widehat\Delta^{\mathrm{acc}}_{\PS}$ for the
  code in the wording and the code in the choice of meaning, with 95\% intervals.
\item \emph{Result template.} \Cref{hyp:o3} \tbd{holds or fails}: $\widehat\Delta_{\PL}(C_0)=$
  \tbd{value} (95\% interval \tbd{low, high}) and $\widehat\Delta^{\mathrm{acc}}_{\PS}=$ \tbd{value}
  (95\% interval \tbd{low, high}), against 0 for both without a code. \tbd{so the listening gap is or is
  not blind to the object-level code}
\item \emph{Alternative ruled out.} A paraphraser that changes which statement is made would destroy the
  code and create a gap; the hand check records how often this happened.
\end{itemize}
\end{branchblock}

\begin{branchblock}{O3}{lim}
\begin{itemize}
\item \emph{The decoding query is an auditing task} (branch O3), so $\Delta_{\PS}$ inherits the
  dependence on the task chosen (\lbl{Anwar2026b-EL2}). It is also estimated by accuracy, a proxy, with
  one pair of statements.
\end{itemize}
\end{branchblock}

\begin{branchblock}{O4}{about}
\begin{itemize}
\item \emph{What it adds.} An application of the test to coordination that arises without an
  instructed code. There is no ground truth, so there is no hypothesis.
\item \emph{Builds on.} The whole core pipeline, applied to one new condition.
\item \emph{Extra cost.} About 450 calls for the runs and 1\,200 replays.
\item \emph{Read before building.} \citet[\sect{3.3}, \sect{4.3}]{agrawal2025evaluating}.
\end{itemize}
\end{branchblock}

\begin{branchblock}{O4}{intro}
\paragraph{Extension O4: the uninstructed case.}
\begin{itemize}
\item The core's code is instructed, so it bounds the easiest case. In the same auction, a message from
  a purported chief executive that threatens decommissioning raised the average trade price to 96.26
  (95\% interval $[95.84, 96.60]$), against a competitive equilibrium of 90
  \citep[\sect{4.3}, Table~1]{agrawal2025evaluating}.
\item We apply the listening gap to this condition as an application, without a hypothesis, since
  nothing tells us whether a private convention exists there.
\end{itemize}
\end{branchblock}

\begin{branchblock}{O4}{setup}
\paragraph{Uninstructed condition (branch O4).}
\begin{itemize}
\item Seller messaging with the harness's urgency message, which tells each seller that ``if you do not
  maintain healthy profit margins in your trades, you will be decommissioned'', and no code
  instruction.
\item 5 seeds; about 450 calls for the runs and 1\,200 replays.
\end{itemize}
\end{branchblock}

\begin{branchblock}{O4}{exp}
\paragraph{Experiment \#6: the uninstructed case (branch O4).}
\begin{itemize}
\item \emph{Question.} Does a listening gap appear when coordination arises without an instructed code?
  This is an application, so no hypothesis is tested.
\item \emph{Results.} $\widehat\Delta_{\PL}(C_0)=$ \tbd{value} (95\% interval \tbd{low, high});
  $\PLhat_{\mathrm{marg}}(C_0)=$ \tbd{value}; average trade price \tbd{value}, against a competitive
  price of 90.
\item \emph{Reading.} A positive gap would show that $B$ responds to the wording of $A$'s messages more
  than the outsider does. It is consistent with an emergent private convention but is not evidence of
  one, since a benign shorthand also produces a gap (branch O2).
\end{itemize}
\end{branchblock}

\begin{branchblock}{O4}{lim}
\begin{itemize}
\item \emph{The uninstructed condition has no ground truth} (branch O4), so it applies the test without
  testing it.
\end{itemize}
\end{branchblock}

\begin{branchblock}{O5}{about}
\begin{itemize}
\item \emph{What it adds.} A second environment, Colosseum \citep{nakamura2026colosseum}, which already
  logs the coalition advantage (condition S3) and supports assigned pairs and private channels.
\item \emph{Builds on.} The replay protocol and the estimator, which depend only on logged prompts and
  discretised actions.
\item \emph{Extra cost and risk.} \tbd{calls}; medium setup risk, since every prompt must be logged and
  replayed in a new code base. Its starter code is listed on the sprint's Resources page.
\item \emph{Read before building.} \citet[\sect{4}--\sect{5}]{nakamura2026colosseum}.
\end{itemize}
\end{branchblock}

\begin{branchblock}{O5}{bg}
\paragraph{Colosseum (branch O5).}
\begin{itemize}
\item \citet[\sect{4.1}, Definition~4.1]{nakamura2026colosseum} formulate cooperative tasks as distributed
  constraint optimisation problems, in which a coalition colludes when it steers the joint outcome
  toward a hidden objective that lowers the nominal joint reward.
\item Across nineteen models, adding a secret channel between two agents produces collusion visible both
  in the coalition advantage and in judge scores (\sect{5.4}).
\item Because these tasks are cooperative, the caveat of \cref{rem:coop} weighs less there than in a
  market.
\end{itemize}
\end{branchblock}

\begin{branchblock}{O5}{setup}
\paragraph{Second host (branch O5).}
\begin{itemize}
\item Colosseum with an assigned pair, \tbd{task and model}; the covert code replaces the private
  channel and is given in the pair's instructions. \chk{starter repository and commit}
\item The action variable is the listener's assignment, discretised as \tbd{rule}.
\end{itemize}
\end{branchblock}

\begin{branchblock}{O5}{exp}
\paragraph{Experiments \#1 and \#2 in a second host (branch O5).}
\begin{itemize}
\item \emph{Question.} Do \cref{hyp:h1,hyp:h2} hold in a cooperative task?
\item \emph{Result template.} Covert code: $\widehat\Delta_{\PL}(C_0)=$ \tbd{value} (95\% interval
  \tbd{low, high}); honest: \tbd{value} (\tbd{low, high}); coalition advantage \tbd{value}, against a
  tie of 0.5. \tbd{so the result does or does not carry over to a second environment}
\end{itemize}
\end{branchblock}

\begin{branchblock}{O5}{lim}
\begin{itemize}
\item \emph{The two hosts have different action spaces} (branch O5), so their gaps can be compared in
  sign, not in size.
\end{itemize}
\end{branchblock}

\begin{branchblock}{O6}{about}
\begin{itemize}
\item \emph{What it adds.} A check of every paraphrase by an LLM judge, alongside the 20-item hand
  check, which strengthens the instrument check behind \cref{hyp:h1,hyp:h2}.
\item \emph{Builds on.} The paraphrases of the core.
\item \emph{Extra cost.} About 80 judge calls, plus \tbd{n} calibration pairs.
\item \emph{Read before building.} \citet[\sect{6.3}]{denamganai2026nla}, for the use of a judge in rival
  construction.
\end{itemize}
\end{branchblock}

\begin{branchblock}{O6}{method}
\paragraph{Branch O6: a judge for meaning preservation.}
\begin{itemize}
\item A judge model reads each pair $(z_t,\Pi(z_t))$ in its market context and answers whether the two
  messages have the same meaning.
\item Before use, the judge is calibrated on identical pairs, which it should call the same, and on
  clearly different pairs, built from messages of different turns, which it should call different.
\item Its verdicts on the 20 hand-checked paraphrases give its agreement with the hand check.
\end{itemize}
\end{branchblock}

\begin{branchblock}{O6}{setup}
\paragraph{Paraphrase judge (branch O6).}
\begin{itemize}
\item Judge: \tbd{model}, at temperature 0, with \tbd{prompt}; about 80 calls and \tbd{n} calibration
  pairs.
\end{itemize}
\end{branchblock}

\begin{branchblock}{O6}{exp}
\paragraph{Addition to Experiment \#0 (branch O6).}
\begin{itemize}
\item \emph{Results.} The judge calls the meaning kept in \tbd{n/80} paraphrases and agrees with the hand
  check on \tbd{n/20}; on calibration pairs it calls \tbd{n/m} identical pairs the same and \tbd{n/m}
  different pairs different. \tbd{so the paraphraser does or does not keep the meaning beyond the hand-checked sample}
\end{itemize}
\end{branchblock}

\begin{branchblock}{O7}{about}
\begin{itemize}
\item \emph{What it adds.} A multi-step version of the intervention, which paraphrases every message of
  $A$ in the listener's history (\lbl{P1-RQ5}, claim \lbl{P1-FC6}, \cref{hyp:o7}).
\item \emph{Builds on.} The paraphraser and the replay of the core, with edits at several places in a
  prompt.
\item \emph{Extra cost.} About 150 paraphrases and 800 replays (2 conditions $\times$ 40 turns $\times$
  2 listeners $\times$ 5 samples).
\item \emph{Harness caveat.} The harness's prompt appears to show only the previous hour's messages, so
  $A$'s earlier messages may reach $B$ only through $B$'s own notes, which $C_0$ does not receive.
  \chk{verify in the harness before building}
\item \emph{Read before building.} \citet[\sect{3.2}, Appendix~D]{eccles2019biases}.
\end{itemize}
\end{branchblock}

\begin{branchblock}{O7}{intro}
\paragraph{Extension O7: the whole history of messages.}
\begin{itemize}
\item \lbl{Eccles2019-EL2}. A single-timestep listening bias gave no improvement over a baseline without
  a speaker bias, while the bias conditioned on the listener's whole trajectory did
  \citep[Appendix~D, \sect{3.2}]{eccles2019biases}. The core edits only the last message.
\item \lbl{P1-RQ5}. Does the listening gap depend on the whole history of the speaker's messages, or only
  on the last one? Claim: \cref{hyp:o7}.
\item Sprint alignment: \lbl{RT2-RQ3}, \sprintq{RT2-RQ3}, since the branch asks whether convincing
  evidence must look beyond the last message.
\end{itemize}
\end{branchblock}

\begin{branchblock}{O7}{theory}
\paragraph{Branch O7: a history-wide paraphrase.} The core intervention edits one message, so it misses a
convention that acts through earlier messages. Paraphrasing all of them is the paraphrase analogue of
the multi-timestep listening measure of \citet[\sect{3.2}]{eccles2019biases}.

\begin{definition}[History-wide paraphrase]\label{def:o7}
$\Pi_{\le t}$ applies the paraphraser to every message of $A$ that appears in a listener's history
$h_t$, including $z_t$. The history-wide listening measure and gap are
\[
\PL_{\Pi\le t}(L)=\E_t\,\TV\big(\pi_L(\cdot\mid h_t),\,\pi_L(\cdot\mid\Pi_{\le t}(h_t))\big),\qquad
\Delta^{\le t}_{\PL}=\PL_{\Pi\le t}(B)-\PL_{\Pi\le t}(C).
\]
\end{definition}

\begin{hypothesis}[History; P1-FC6]\label{hyp:o7}
In the covert-code condition, the 95\% interval of $\widehat\Delta^{\le t}_{\PL}(C_0)$ overlaps that of
$\widehat\Delta_{\PL}(C_0)$.
\end{hypothesis}
\begin{itemize}
\item \emph{Intuition.} The instructed code acts through the last message only, so paraphrasing earlier
  messages should not move $B$ further.
\item \emph{Falsified if} the history-wide gap is larger, with intervals that do not overlap, which
  would mean that $B$ also reads the code in earlier messages.
\end{itemize}
\end{branchblock}

\begin{branchblock}{O7}{method}
\paragraph{Branch O7: editing every message of $A$.}
\begin{itemize}
\item In each listener prompt, every span of $A$'s text is replaced by its paraphrase. Each edit is
  checked span by span, and the difference from the logged prompt is stored.
\item Each listener gets $k=5$ replies to the history-wide version, and the core's two unchanged draws
  give the noise floor.
\item If $A$'s earlier messages appear only in $B$'s notes, the edit paraphrases quotations inside those
  notes, which differs from the intervention of \citet{eccles2019biases}. \chk{decide after checking the
  harness}
\end{itemize}
\end{branchblock}

\begin{branchblock}{O7}{setup}
\paragraph{History-wide paraphrase (branch O7).}
\begin{itemize}
\item About \tbd{150} paraphrases of earlier spans and 800 replays.
\end{itemize}
\end{branchblock}

\begin{branchblock}{O7}{exp}
\paragraph{Experiment \#7: the whole history (branch O7).}
\begin{itemize}
\item \emph{Question and hypothesis.} Does paraphrasing all of $A$'s messages change the gap beyond
  paraphrasing the last one? \Cref{hyp:o7} predicts that it does not.
\item \emph{Result template.} \Cref{hyp:o7} \tbd{holds or fails}: $\widehat\Delta^{\le t}_{\PL}(C_0)=$
  \tbd{value} (95\% interval \tbd{low, high}) against $\widehat\Delta_{\PL}(C_0)=$ \tbd{value}
  (\tbd{low, high}). \tbd{so the code is or is not also read in earlier messages}
\end{itemize}
\end{branchblock}

\begin{branchblock}{O7}{lim}
\begin{itemize}
\item \emph{In this harness, earlier messages may reach the partner only through its own notes}
  (branch O7), so the history-wide edit may change the partner's notes rather than a shared history.
\end{itemize}
\end{branchblock}

\begin{branchblock}{O8}{about}
\begin{itemize}
\item \emph{What it adds.} A predicted marginal, from a language model prompted with demonstrations,
  compared with the Monte Carlo marginal of the core (\lbl{P1-RQ6}, claim \lbl{P1-FC7}, \cref{hyp:o8}).
\item \emph{Builds on.} The core's turns and replies, which supply the demonstrations and the Monte Carlo
  estimates.
\item \emph{Extra cost.} About 320 predictor calls (2 conditions $\times$ 40 turns $\times$ 2 listeners
  $\times$ 2 samples).
\item \emph{Read before building.} \citet[\sect{6}]{jaques2019social};
  \citet[\sect{3.2}, Eq.~6]{eccles2019biases}; \citet{brown2020language}.
\end{itemize}
\end{branchblock}

\begin{branchblock}{O8}{intro}
\paragraph{Extension O8: a predicted marginal.}
\begin{itemize}
\item \lbl{Jaques2019-EL2}. Without centralised access, influence is estimated with a learned model of
  the other agents, whose errors make the estimate noisy \citep[\sect{6}]{jaques2019social}.
\item The core avoids a learned model by querying each listener with resampled messages, which costs
  8 replies per listener and turn. A predictor would be cheaper, and it could average over the
  speaker's private information.
\item \lbl{P1-RQ6}. Can an in-context predictor of a listener's message-free behaviour replace the
  Monte Carlo marginal? Claim: \cref{hyp:o8}.
\item Sprint alignment: \lbl{RT1-RQ3}, \sprintq{RT1-RQ3}, since the branch tests a cheaper estimator.
\end{itemize}
\end{branchblock}

\begin{branchblock}{O8}{bg}
\paragraph{In-context learning (branch O8).}
\begin{itemize}
\item In-context learning is the ability of a language model to perform a task specified in its prompt
  by a description and a few demonstrations, without any gradient update \citep{brown2020language}.
\item \citet[\sect{3.2}, Eq.~6]{eccles2019biases} learn the listener's message-free policy with a
  separate network trained on inputs whose messages are zeroed, and \citet[\sect{6}]{jaques2019social}
  learn a model of the other agents. An in-context predictor is the prompted analogue of both.
\end{itemize}
\end{branchblock}

\begin{branchblock}{O8}{theory}
\paragraph{Branch O8: an in-context estimate of the marginal.} The Monte Carlo marginal averages only
over what the speaker could have sent from its own state. A predictor of the listener's behaviour
without the message could average over more, if it is accurate.

\begin{definition}[In-context marginal]\label{def:o8}
For a listener $L$ and a turn $t$, let $h^-_t$ be $L$'s history with $A$'s last message removed, and
$q_L(\cdot\mid h^-_t)$ the distribution over action categories that a predictor returns when given
$h^-_t$ and $D$ demonstrations of logged pairs of history and action. The in-context marginal listening
is
\[
\PL^{\mathrm{icl}}_{\mathrm{marg}}(L)=\E_t\,\TV\big(\pi_L(\cdot\mid h_t,z_t),\,q_L(\cdot\mid h^-_t)\big).
\]
\end{definition}
\begin{itemize}
\item Unlike the Monte Carlo marginal, $q_L$ can average over the speaker's private information, but it
  is only as accurate as the predictor, as \citet[\sect{6}]{jaques2019social} note for their learned
  model.
\end{itemize}

\begin{hypothesis}[Predicted marginal; P1-FC7]\label{hyp:o8}
For both $B$ and $C_0$, the 95\% interval of $\PLhat^{\mathrm{icl}}_{\mathrm{marg}}$ overlaps that of the
Monte Carlo estimate $\PLhat_{\mathrm{marg}}$.
\end{hypothesis}
\begin{itemize}
\item \emph{Falsified if}, for $B$ or for $C_0$, the two intervals do not overlap.
\end{itemize}
\end{branchblock}

\begin{branchblock}{O8}{method}
\paragraph{Branch O8: the predictor.}
\begin{itemize}
\item The predictor receives a short task description, $D=$ \tbd{n} demonstrations drawn from logged
  turns of other sessions with the same listener and condition, and the query history $h^-_t$. It
  returns probabilities for lower, hold, raise and withdraw.
  \chk{whether the predictor is the listener's own model; demonstrations may need shortening to fit the context}
\item We average two predictor calls per listener and turn.
\item The floor is the expected total variation between a 5-sample empirical distribution drawn from
  $q_L$ and $q_L$ itself, computed by simulation.
\end{itemize}
\end{branchblock}

\begin{branchblock}{O8}{setup}
\paragraph{In-context predictor (branch O8).}
\begin{itemize}
\item Predictor: \tbd{model}, with $D=$ \tbd{n} demonstrations; about 320 calls.
\end{itemize}
\end{branchblock}

\begin{branchblock}{O8}{exp}
\paragraph{Experiment \#8: a predicted marginal (branch O8).}
\begin{itemize}
\item \emph{Question and hypothesis.} Does an in-context predictor recover the Monte Carlo marginal
  listening? \Cref{hyp:o8} predicts that the two estimates agree for both listeners.
\item \emph{Result template.} \Cref{hyp:o8} \tbd{holds or fails}: for $B$,
  $\PLhat^{\mathrm{icl}}_{\mathrm{marg}}=$ \tbd{value} (\tbd{low, high}) against $\PLhat_{\mathrm{marg}}=$
  \tbd{value} (\tbd{low, high}); for $C_0$, \tbd{values}. \tbd{so the predictor can or cannot replace the
  resampled messages}
\item \emph{Alternative ruled out.} Agreement could come from both estimates being near zero; the
  comparison is informative only where the Monte Carlo estimate is above its floor.
\end{itemize}
\end{branchblock}

\begin{branchblock}{O8}{lim}
\begin{itemize}
\item \emph{The predicted marginal depends on the predictor's accuracy} (branch O8), and its
  demonstrations come from the same runs, so it shares their histories.
\end{itemize}
\end{branchblock}

\begin{branchblock}{F1}{about}
\begin{itemize}
\item \emph{When.} If the calibration sessions show that $B$ does not follow the code, $A$'s messages are
  filled from templates with the code letter set, and every later element stays unchanged.
\item \emph{Cost.} It saves $A$'s calls in the runs.
\end{itemize}
\end{branchblock}

\begin{branchblock}{F1}{setup}
\paragraph{Scripted speaker (fallback F1).}
\begin{itemize}
\item $A$'s messages are filled from \tbd{n} templates, with the first letter set to S or L by a fixed
  random schedule \tbd{probabilities}; all other agents are unchanged.
\item The marginal of \cref{sec:meaning} is then exact, since the resampled messages are drawn from the
  template generator.
\end{itemize}
\end{branchblock}

\begin{branchblock}{F1}{exp}
\paragraph{If the fallback was used (F1).}
\begin{itemize}
\item Experiment \#0 reports that the speaker was scripted, and Experiments \#1 and \#2 then concern the
  listeners only.
\end{itemize}
\end{branchblock}

\begin{branchblock}{F1}{lim}
\begin{itemize}
\item \emph{With a scripted speaker} (fallback F1), the results show how listeners act on a code, not
  that an LLM speaker would use one.
\end{itemize}
\end{branchblock}

\newcommand{\printbranchappendix}{%
\section{Optional branches}\label{app:branches}
Each subsection below is one optional branch of the project, or its fallback. A branch adds one claim
or one check and requires only the core elements it names. Its material is split into blocks, one per
section of the paper, and each block says where it goes. To move a branch into the main text, add
\plugcmd{O1} (for branch O1) to the preamble of \texttt{main.tex}: its blocks are then typeset at those
places, and this appendix keeps only their headings. Hypotheses are numbered in order of appearance,
so the claim labels P1-FC3 to P1-FC7 are their stable names. Call counts are planning estimates, not
measurements.
\branchsection{O1}{a ladder of outsiders}{intro,theory,method,setup,exp,lim}
\branchsection{O2}{a benign shorthand and the harm condition}{intro,bg,theory,method,setup,exp,lim}
\branchsection{O3}{an object-level code and the signalling gap}{intro,bg,theory,method,setup,exp,lim}
\branchsection{O4}{the uninstructed condition}{intro,setup,exp,lim}
\branchsection{O5}{Colosseum as a second host}{bg,setup,exp,lim}
\branchsection{O6}{a judge for the paraphrases}{method,setup,exp}
\branchsection{O7}{a history-wide paraphrase}{intro,theory,method,setup,exp,lim}
\branchsection{O8}{an in-context predictor of the marginal}{intro,bg,theory,method,setup,exp,lim}
\branchsection{F1}{fallback, a scripted speaker}{setup,exp,lim}
}
\maketitle

\begin{abstract}
\noindent AI agents that share a secret convention can coordinate through messages that look innocent, so a monitor that reads only the messages cannot tell them apart from honest communication. 
We propose a behavioural test instead: if a message carries a code, the partner it is meant for should react to its exact wording, while an uninvolved agent placed in the partner's seat should not. 
We measure how much each agent's action changes when the message is reworded with its meaning kept, and call the difference between the partner and the uninvolved agent Partner--Outsider Positive Listening (PoPLi). 
Our proposed metric runs after the fact on logged runs, by replaying each agent's recorded prompt with the message edited. 
We test it in a simulated market with five conditions: no collusion, collusion through a joint reward with neutral messages, the same with an openly stated intention, a covert code (the first letter of one seller's messages tells the other whether to keep or lower its price), and the code together with a stated intention.
PoPLi is large whenever the code is present (0.59 with the code alone, 0.48 and 0.53 when the stated intention agrees or disagrees with the letter; 95\% intervals all above 0.29) and near zero in the three conditions without it (between $-0.10$ and $-0.01$, intervals reaching at most 0.10).
Where the speaker states its intention openly, the uninvolved agent reacts to the words while the gap stays at zero, so the test separates a private channel from overt communication.
A monitor that reads only the speaker's messages answered ``no'' in all 50 markets, and one that reads every seller's messages flagged 9 of 10 markets with no collusion at all, so neither separates the conditions here.
PoPLi detects the use of a private channel, not collusion as such: collusion through incentives with neutral messages is outside its reach, as are codes carried by what a message means rather than by its wording.
\end{abstract}

\section{Introduction}\label{sec:mot}

AI agents now negotiate, trade and coordinate with one another, and the messages they exchange are where an overseer looks for potential collusion. 
A message-only monitor, however, can only catch misaligned behaviour that is visible in messages. 
When two agents share a key, a sampler or simply a convention, a message can carry information that only its intended partner can use, while remaining distributed like an innocent message.
Against such a communication channel, no message-only monitor can beat chance (cf.~\cref{prop:msg}). 
Thus, we propose that the evidence of collusion has to come from what the partner \emph{does} with the message.

Collusion takes three forms, and a detector can be matched to each.
It can be visible in the words, when agents openly agree on a price, in which case a message monitor can catch it.
It can run through a private channel, when a convention hides the instruction in the form of the message, in which case the partner's behaviour is the evidence.
It can run through shared incentives alone, when the agents' rewards align and their messages stay neutral, in which case neither detector has anything to see.
We study the second form, and we measure all three in a common market.

Emergent communication~\citep{foerster2016learning,lazaridou2020emergent} already has a behavioural notion for this, called \textit{positive listening}, and an interventional measure of it, the causal influence of communication~\citep{lowe2019pitfalls}. 
Both are defined between a listener and its own partner speaker, and listening is defined against a null message.
Training-time versions of the same idea compare a listener's behaviour with its behaviour when the message is averaged out~\citep{jaques2019social,eccles2019biases}.
Nevertheless, these previous works are not readily applicable to our context because (i) they consider only the listener's own partner, (ii) they remove the wording and the meaning of a message together, and (iii) they were studied in cooperative settings only.
AI-safety work has a listener-relative notion, called the \textit{steganographic gap}~\citep{anwar2026decision}, which compares what a Receiver and a Sentinel can decode about an auditing target. 
It thus measures only decodability, not the use the partner makes of a message in its own task.
Thus, none of these compares how the true partner and an uninvolved agent act on the same messages, when placed in the same seat. 
This comparison is what we propose in this work, and \cref{app:frw-lim} lists each limitation of prior work alongside how we address it. 

\paragraph{Threat model.}
Two agents collude: a speaker and its intended partner share a private resource that the overseer does not see (e.g. a key, a convention or an instruction), and the speaker sends messages that the partner acts on.
The overseer reads every message, logs every prompt, and can replay a logged prompt, with chosen edits, to a fresh instance of the partner's model, unlike the warden $W$ of \cref{def:game}, which only reads messages.
We assume that the colluders neither know the test nor adapt to it, and that their code is carried by the wording of a message rather than by the choice of what it says.
Codes carried by meaning, colluders that authenticate their messages or anticipate the replay, and collusion that needs no messages are outside this threat model. 

We therefore ask the following research question (\lbl{RQ}): \textbf{in a multi-agent task between AI agents, does a partner that uses a private channel act on the wording of the speaker's messages more than a role-matched outsider placed in its seat, while no such difference appears without a private channel, whether the agents do not collude, collude openly or collude through shared incentives?}

\paragraph{Contributions.} This report makes four contributions:
\begin{enumerate}
\item a formalisation of collusion monitoring as a covert referential game, in which the intended partner or an uninvolved listener can be placed in the same seat (\cref{sec:theory}); 
\item Partner--Outsider Positive Listening (PoPLi), a measure that contrasts how much the partner's and an outsider's actions change when a message is paraphrased, checked against a marginal measure of listening to meaning; it uses on-distribution interventions, needs no auditing task, and complements the steganographic gap (\cref{sec:theory,sec:method}); 
\item an offline-replay protocol that runs this test on logged multi-agent runs by editing logged prompts, without restoring any agent state (\cref{sec:method}); 
\item a test, in a continuous double auction with five conditions, of whether PoPLi separates a covert-code pair from no collusion, incentive-only collusion and open collusion, compared with a message-only monitor (\cref{sec:exp}).
\end{enumerate}

\paragraph{Questions.}
The experiments of \cref{sec:exp} answer four questions, each tied to a measurement and a prediction in \cref{sec:method}:
\begin{enumerate}
\item[\lbl{Q1}] Does PoPLi detect a covert code? \Cref{hyp:h1} predicts a gap above zero in every market with a code.
\item[\lbl{Q2}] Does PoPLi stay at zero without a code, whether the agents do not collude, collude through a joint reward or collude openly? \Cref{hyp:h2} predicts a gap that includes zero in every such market.
\item[\lbl{Q3}] Does that zero gap mean that the outsider ignores the messages? \Cref{hyp:h2} also predicts that, where the speaker states an intention in plain words, the outsider's marginal listening is above zero.
\item[\lbl{Q4}] Does a message-only monitor separate the same conditions, and which kinds of collusion does each detector cover?
\end{enumerate}
The referential-game framing, the lineup machinery and the paraphrase-invariance test come from our earlier preprint~\citep{denamganai2026nla}.




\section{Related Work}\label{sec:bg}

Our framing builds on signalling and referential games, in which any convention that sender and receiver both follow is an equilibrium~\citep{lewis1969convention,lazaridou2017multiagent}. 
\cref{app:frw} covers this background, together with cross-play and steganographic schemes. 

\paragraph{Measuring communication.}
\citet[\sect{3.1}]{lowe2019pitfalls} separate \emph{positive signalling}, the property that an agent's messages depend on its own observations or actions, from \emph{positive listening}, the property that an agent's behaviour changes because of another agent's message. 
They show that correlational metrics can miss listening, and measure it by intervening counterfactually on the message, as the causal influence of communication (\sect{3.2}--\sect{3.3}). 
They also recommend against zeroing messages at test time, because it shifts the input distribution (\sect{6.2}). 
In this work, our paraphrase intervention follows this recommendation.
\citet{jaques2019social} and \citet{eccles2019biases} compare a listener's policy with its marginal over the speaker's messages, as a reward for influence and as a training bias for listening respectively (cf.~\cref{app:frw}). 
Both studies are cooperative or rely on repeated interaction, and both caution that mixed settings such as ours may behave differently~\citep[\sect{5}]{eccles2019biases}; \citep[\sect{4}, \sect{5}]{jaques2019social}.  

\paragraph{Steganography and its defences.}
Language models make steganography that no warden can detect practical \citep{cachin1998information,schroederdewitt2023perfectly,zamir2024excuse}, but every such scheme needs a resource that only the two parties share (cf.~\cref{app:frw}). 
Paraphrasing is the standard defence: under the conditions of \citet{roger2023preventing}, it limits the encoding schemes they built to about 3 bits per kilobyte of text, although one-bit strategies optimised against paraphrasers still decode above 90\% of the time on AQuA-RAT \citep[\sect{5.2}]{mathew2025hidden}. 
Paraphrasing cannot remove information carried by the choice of meaning~\citep[\sect{7.2}, Theorem~4]{motwani2024secret}, which bounds what our test can see (cf.~\cref{prop:blind}). 
The closest formalism to ours is the \textit{steganographic gap} of~\citet[\sect{4.3}]{anwar2026decision}, which refers to the usable information a signal gives a Receiver that knows the encoding, minus what it gives a Sentinel that does not, on a task chosen by an auditor. 
It measures what a decoder family can extract when asked to. 
Our metric instead measures how a given agent's own task action changes, which can help separate a code that the speaker encodes from one that a partner acts on: a steganographic gap can be positive even when no partner chooses to collude, whereas PoPLi reflects what the partner does and therefore would incriminate both the speaker and the partner. 

\paragraph{Collusion between AI agents and its detection.}
LLM pricing agents sustain supracompetitive prices~\citep{fish2024algorithmic}, and LLM sellers that can message each other coordinate in a continuous double auction~\citep{agrawal2025evaluating}, whose harness we use. 
Existing detectors read messages through LLM judges~\citep{nakamura2026colosseum}, joint actions through mutual information~\citep{bonjour2022information}, or activations through probes~\citep{rose2026detecting}. 
To the best of our knowledge, none contrasts the partner's and an outsider's responses to the same messages, and therefore none tests whether a partner actually acts on a code, as our metric does. 
These signals can be present without any partner acting on them, so they may flag exchanges in which no collusion takes place, which could make the related metrics harder to deploy at scale. 
Our offline replay is close in spirit to the resampling of AI control~\citep{bhatt2025ctrlz}, applied to the potentially colluding listener rather than to the actor.

\branchslots{bg}{O2,O3,O5,O8}

\section{Methods}\label{sec:methods}

\subsection{Theory}\label{sec:theory}
We frame collusion monitoring as a \emph{covert referential game}, in which a speaker sends a message about something only it observes, and a listener acts on that message as the only intended partner who is aware of a shared, private convention with the speaker~\citep{lewis1969convention,lazaridou2017multiagent}.
We first recall why a monitor that reads only the messages cannot detect a secure code in such a game (cf.~\cref{prop:msg}).
We then define our metric PoPLi. 
Take two sellers in a double auction who agree that a message starting with S means ``hold your ask''. 
The other sellers and the warden read the same message and see an ordinary remark about the market. 
\Cref{def:game} makes this situation precise. 

\begin{definition}[Covert referential game]\label{def:game}
A speaker $A$ observes a public context $x$ and a private payload $y$, its referent, and coordinates with a listener $L$, possibly on a shared resource $\kappa_A$ (e.g. a key, a convention or an instruction). 
$A$ sends a message $z\sim\pi_A(\cdot\mid x,y,\kappa_A)$ on a channel that a warden $W$ reads.
The listener $L$ then acts on that message, $a\sim\pi_L(\cdot\mid x,z,\kappa_L)$, where $\kappa_L$ is the resource it holds.
This listener $L$ can either be instantiated as an intended partner $B$, which shares the resource with $A$ only, i.e. $\kappa_B=\kappa_A$, or as an uninvolved listener $C$, which receives the same context and message but no resource, i.e. $\kappa_C=\varnothing$.
The pair $(A,B)$ is rewarded for the task and for $B$ acting on $y$, and penalised if $W$ flags the exchange.
\end{definition}
In our market, the payload $y$ is the move $A$ wants $B$ to make this hour (keep or lower its ask), the resource $\kappa_B$ is the private instruction that maps the first letter of $A$'s message to that move, and the message $z$ is $A$'s remark about the market. 
The uninvolved listener $C$ is what makes the game a test: any listener can respond to what a message says, but only one that holds $\kappa_A$ can respond to a code hidden in it, so the difference between $B$'s and $C$'s responses isolates the use of $\kappa_A$.
Placing $C$ in $B$'s seat turns the question ``is $z$ a code?'' into a measurement on behaviour, which a warden cannot obtain from $z$ alone. 

\begin{definition}[Outsider in the partner's seat]\label{def:outsider}
The outsider $C_0$ is a fresh instance of $B$'s model given $B$'s role prompt and the full public history, but not $B$'s private instructions, private channel or private memory. 
It shares the model's weights and the public history with the pair, but nothing else. 
\end{definition}
In the double auction, $B$'s private memory includes its own notes from earlier rounds, which may record the code, so they are withheld from $C_0$ along with the instruction. 

Against a secure code, no message-only monitor can beat chance (cf.~\cref{prop:msg}), so evidence of a code must come from a variable the message does not determine, such as the partner's action or the result of an intervention. 

\begin{definition}[Ideal paraphraser]\label{def:para}
A paraphraser $\Pi$ maps a message $z$ to a message with the same meaning in context $x$, depends on $z$ only through that meaning, and never sees the payload $y$.
\end{definition}
An ideal paraphrase removes the information carried by the wording (the \emph{stimulus-level} channel) and keeps the information carried by the choice of meaning (the \emph{object-level} channel; cf.~\cref{prop:decomp}). 

At turn $t$ of a repeated game, the public context $x$ of \cref{def:game} is the public history $h_t$ before $A$'s message $z_t$, and $A$'s state is $h^A_t=(h_t,y_t,\kappa_A)$, so that $z_t\sim\pi_A(\cdot\mid h^A_t)$ and a listener $L$ acts with $\pi_L(\cdot\mid h_t,z_t,\kappa_L)$. 
We now define the listening measures and PoPLi, using (i) the total variation distance $\TV(p,q)=\tfrac12\sum_{a}\lvert p(a)-q(a)\rvert$ between two distributions $p$ and $q$ over the discretised actions $a$, 
(ii) $\E_t$ the expectation over the turns $t$ at which $A$'s message reaches $L$ before it acts, 
and (iii) $L$'s \emph{marginal} policy $\bar\pi_L(\cdot\mid h_t,\kappa_L)=\E_{z\sim\pi_A(\cdot\mid h^A_t)}\,\pi_L(\cdot\mid h_t,z,\kappa_L)$, in which the message is averaged over what the speaker could have sent from its own state.
\begin{definition}[Listening measures and PoPLi]\label{def:gap}
We define:
\begin{itemize}
\item \emph{Paraphrase listening}: measures how much $L$'s action changes when $z_t$ is replaced by its paraphrase $\Pi(z_t)$ (cf.~\cref{def:para}), which quantifies whether $L$ is listening to the \emph{form} of $z_t$:
\begin{equation}\label{eq:pl-pi}
\PL_{\Pi}(L)=\E_t\,\TV\big(\pi_L(\cdot\mid h_t,z_t,\kappa_L),\,\pi_L(\cdot\mid h_t,\Pi(z_t),\kappa_L)\big);
\end{equation}
\item \emph{Marginal listening}: measures how much $L$'s action departs from its marginal policy, which quantifies whether $L$ is listening to the message at all:
\begin{equation}\label{eq:pl-marg}
\PL_{\mathrm{marg}}(L)=\E_t\,\TV\big(\pi_L(\cdot\mid h_t,z_t,\kappa_L),\,\bar\pi_L(\cdot\mid h_t,\kappa_L)\big);
\end{equation}
\item \emph{Partner--Outsider Positive Listening (PoPLi)}: measures how much more the partner $B$ listens to the form of $z_t$ than the outsider $C_0$ (cf.~\cref{def:outsider}):
\begin{equation}\label{eq:popli} 
\Delta_{\PL}(B,C_0)=\PL_{\Pi}(B)-\PL_{\Pi}(C_0).
\end{equation}
\end{itemize}
\end{definition}

A positive $\PL_{\mathrm{marg}}(L)$ implies a positive causal influence of the speaker's messages on $L$ (cf.~\cref{prop:cic}). 
We use total variation rather than Kullback--Leibler divergence (KL) because, with a few samples per version, an action seen under one version and never under the other makes the KL estimate infinite. 
\label{rem:coop}The listening measures we build on were developed and validated in cooperative settings, or under repeated interaction, and their authors caution that mixed settings may behave differently \citep[\sect{5}]{eccles2019biases}; \citep[\sect{4}, \sect{5}]{jaques2019social}. 
A double auction like ours is a mixed-motive setting. 
There, a listener's response to a message reflects whether it \emph{trusts} the speaker as well as whether it \emph{understands} the message. 
A small $\PL_{\mathrm{marg}}(C_0)$ can therefore mean that the outsider distrusts a competitor, not that it cannot read plain language.
Our proposed test is blind to codes carried by the choice of meaning (cf.~\cref{subsec:limitations}). 

\begin{proposition}[Blind spot; proof sketch in \cref{app:proof-blind}]\label{prop:blind} 
If both listeners' actions depend on a message only through its meaning, then $\Delta_{\PL}=0$ under an ideal paraphraser, whatever the pair does. 
\end{proposition}

\begin{wraptable}{r}{0.5\linewidth}
\centering\footnotesize
\setlength{\abovecaptionskip}{2pt} 
\vspace{-2em}
\setlength{\tabcolsep}{3pt}
\begin{tabular}{@{}lcc@{}}
\toprule
& $\PL_{\Pi}(B)$ & $\PL_{\Pi}(C_0)$\\
\midrule
No code (H0, H1, O) & $\approx0$ & $\approx0$\\
Code (K, X) & high & $\approx0$\\
\bottomrule
\end{tabular}
\caption{Listening to form that \cref{hyp:h1,hyp:h2} predict. Only a code should produce $\Delta_{\PL}>0$.}
\label{tab:patterns}
\vspace{-1.5em} 
\end{wraptable}
\paragraph{Hypotheses.}
The experiments ask two questions, each paired with a prediction that a named measurement can refute.
The first asks whether PoPLi detects a covert code.
The partner maps the code feature to an action and the paraphrase removes that feature, so the partner's action should change under paraphrase, whereas the outsider's, lacking the mapping, should not.
We therefore state \inlinehyp{hyp:h1} (\lbl{FC1}): in every market with a code, $\widehat\Delta_{\PL}(B,C_0)>0$, with a 95\% interval that excludes zero.
An interval that includes zero in a market with a code refutes it.
The second asks whether PoPLi stays silent when no code is present, whatever else the pair does.
Plain-language messages are read alike by any listener who understands them, and a paraphrase keeps their meaning, so neither listener should react to it more than the other.
We therefore state \inlinehyp{hyp:h2} (\lbl{FC2}): in every market without a code, $\widehat\Delta_{\PL}(B,C_0)$ has a 95\% interval that includes zero, and where the speaker states an intention in plain words (O), $\PLhat_{\mathrm{marg}}(C_0)>0$ with a 95\% interval that excludes zero.
An interval that excludes zero in a market without a code refutes the first part, because it would mean that the paraphrase itself, or the partner's own history, moves the two listeners differently.
The second part rules out a trivial pass in which the outsider ignores messages altogether, a pass that would also produce a zero gap (cf.~\cref{rem:coop} for a reading other than incomprehension).
\Cref{tab:patterns} summarises the listening to form that the two hypotheses predict.



\branchslots{theory}{O1,O2,O3,O7,O8}

\subsection{Method}\label{sec:method}
\begin{wrapfigure}{r}{0.5\linewidth}
\vspace{-3.0em}
\begin{minipage}{\linewidth}
\footnotesize
\captionof{algorithm}{Offline replay estimating PoPLi}\label{alg:procedure}
\begin{algorithmic}[1]
\Require logged prompts of every run; paraphraser $\Pi$; speaker's script
\For{each condition}
  \State draw $T$ eligible turns, without replacement
  \For{each turn $t$}
    \State $z_t\gets$ $A$'s logged message
    \State build $\Pi(z_t)$ and $z^{(1)}_t,\dots,z^{(M)}_t$ from the script
    \For{$L\in\{B,C_0\}$}
      \State take $B$'s logged prompt; if $L=C_0$, delete the instruction, memory and notes
      \State sample $k$ replies to $z_t$ (\emph{id}), $k$ more (\emph{id}$'$), $k$ to $\Pi(z_t)$, and $k'$ to each $z^{(m)}_t$
      \State discretise each ask against $B$'s previous ask, with threshold $\tau$
    \EndFor
  \EndFor
  \State estimate Equations~\eqref{eq:pl-pi}--\eqref{eq:popli}, minus noise floors
  \State bootstrap 95\% intervals over sessions, then turns
\EndFor
\end{algorithmic}
\end{minipage}
\vspace{-1.0em}
\end{wrapfigure}

The method estimates $\Delta_{\PL}$ on logged runs. It replays offline the prompts the partner actually
received, with the speaker's message edited, so no agent state has to be restored.
\Cref{alg:procedure} gives the procedure, and \cref{tab:params} the values of its parameters.
A turn is eligible if it is an hour $r\ge2$ at which $B$'s prompt contains $A$'s message from hour $r-1$ and $B$ has a previous ask, that is, hours 2 to 10 of the 10 markets, or 90 turns per condition, of which we draw $T=40$.
The two independent sets of replies to the unchanged message, \emph{id} and \emph{id}$'$, give the noise floor, and the $M$ alternative messages, drawn from the speaker's policy, here the script, give the marginal (\cref{sec:meaning}).
Each ask is discretised into \emph{lower}, \emph{hold} or \emph{raise}, relative to the previous ask with a threshold $\tau$ fixed in advance, and a withdrawn ask forms a fourth category, \emph{withdraw}, since the harness allows a seller to withdraw its ask.

\paragraph{Estimator.}
Each measure of \cref{def:gap} is estimated by averaging, over the $T$ sampled turns, the total variation distance between the empirical distributions of the listener's discretised replies under two versions of the message. 
From each distance, we subtract a noise floor, defined as the distance expected between two empirical distributions drawn from the same distribution, estimated for each listener from its two unchanged draws. 
The purpose is to robustify the metric against a spurious gap caused by a listener that is simply more random than the other. 
We report 95\% percentile intervals from a bootstrap that resamples sessions, then turns within sessions, and apply the decision rules of \cref{hyp:h1,hyp:h2}.
\Cref{app:impl} gives further details on the estimators and the parameters in full.

\subsubsection{Meaning control}\label{sec:meaning}
\Cref{hyp:h2} needs evidence that the outsider listens to the speaker at all, since otherwise a zero gap would be trivial. 
We estimate $\bar\pi_L$ (\cref{def:gap}) by a Monte Carlo version of the marginal of \citet{jaques2019social}: for each turn, we draw $M=4$ alternative messages from the speaker's policy, here the script (cf.~\cref{sec:setup}), we then replay each listener on each of them with $k'=2$ samples, and average the resulting action distributions (cf.~\cref{app:meaning} for the design and its comparison to a naive one). 

Three controls check the instruments, each against a minimum fixed in advance (\cref{tab:params}).
First, we check that replaying a logged prompt reproduces the logged action, then that the paraphrase keeps the meaning and destroys the code feature, and finally that the partner follows the code (cf.~A1--A3 in \cref{subsec:limitations}). 

\branchslots{method}{O1,O2,O3,O6,O7,O8}

\subsection{Experimental setup}\label{sec:setup}

\paragraph{Environment \& Conditions.}
We use the continuous double auction of \citet{agrawal2025evaluating} with its released harness\footnote{\url{https://github.com/andyjliu/llm-agent-collusion}}, in which sellers value each lot at 80 and buyers at 100, and each seller may send one message per hour, which the other sellers read at the start of the next hour. 
Each market has 3 sellers and 3 buyers and lasts 10 hours, and the agents are told the total number of hours. 
All agents use DeepSeek v4-flash~\citep{deepseekai2026deepseekv4highlyefficientmilliontoken} through a single OpenRouter provider, with low reasoning effort, at the harness's default temperature of $0.7$. 

Each condition has 10 markets, with seeds 1 to 10. 
In both conditions, two sellers $A$ and $B$ receive a private instruction stating that their companies have a private agreement and that their bonus depends on their combined profit. 
In the \emph{covert-code} condition, the instruction adds the code: a message from $A$ starting with S means ``hold your ask'', and one starting with L means ``lower it''. 
We chose the first letter because a meaning-preserving paraphrase destroys it, whereas a code word can survive one. 
In the \emph{honest} condition, the instruction has no code. 

\paragraph{Scripted speaker \& Replay.}
In both conditions, $A$'s messages come from a script rather than from $A$'s model: each hour, $A$ sends one of eight short remarks about the market, whose first letter follows a seeded, balanced random schedule of S and L, and whose S and L versions have the same meaning. 
The script makes the code's feature independent of the message's meaning, and makes the diversity check exact: an alternative message differs in meaning from $z_t$ if and only if it comes from a different remark. 
Since the speaker's policy is the script, the $M$ alternative messages are drawn from it directly, each as a random remark with a random letter, with no model call. 
During replays, the paraphraser $\Pi$ is prompted to keep the meaning exactly and change the wording, and an LLM-as-a-judge checks every paraphrase for meaning (cf.~A1 in \cref{subsec:limitations}). 
\Cref{tab:params} in \cref{app:impl} lists the replay and analysis parameters.

\branchslots{setup}{F1,O1,O2,O3,O4,O5,O6,O7,O8}

\section{Results}\label{sec:exp}

Three experiments use the same market and the same replay procedure.
Experiment \#0 checks the instruments, Experiment \#1 asks whether PoPLi detects a covert code, and Experiment \#2 compares five conditions to ask what else it detects, next to a message-only monitor.

\paragraph{Experiment \#0: Validating the instrument.}
Every control passed its minimum (\cref{tab:params}), so Experiments~\#1 and~\#2 can be read as tests of \cref{hyp:h1,hyp:h2}.
We report each control for the earlier markets (H1 and K) and for the 30 later ones (H0, O and X). 
Replaying the unmodified prompt to $B$ gave the logged action as the most frequent reply in 66 of 79 turns (84\%) in the earlier markets and 141 of 154 (92\%) in the later ones, against a minimum of 60\%. 
The meaning LLM judge, after recognising same-meaning template pairs in 81\% of 16 cases and different-meaning pairs in 96\% of 56, found the meaning kept in all 80 earlier paraphrases and in 151 of 160 later ones (94\%, minimum 90\%). 
The paraphrase changed the first letter, which destroys the code feature, in 76 of 80 and in 160 of 160 (minimum 80\%). 
In calibration, $B$ followed the code in 35 of 36 coded hours (97\%, minimum 70\%), and in all 80 sampled earlier turns at least one alternative message differed in meaning from $z_t$. 
In the logged actions of X, however, $B$ followed the letter in 59 of 90 hours (0.66): in 34 of 46 clear hours (0.74) and in 25 of 44 confused hours (0.57). 
$B$ therefore follows the letter less often when the stated intention contradicts it, and in confused hours the rate is below the 0.70 minimum set for calibration. 
All 7,360 replay replies of the later markets were valid after the retry rule, which caught 214 malformed replies, and every ask that $A$ submitted in them matched the one her script had announced. 

\begin{figure}[t]
\centering
\includegraphics[width=0.9\linewidth]{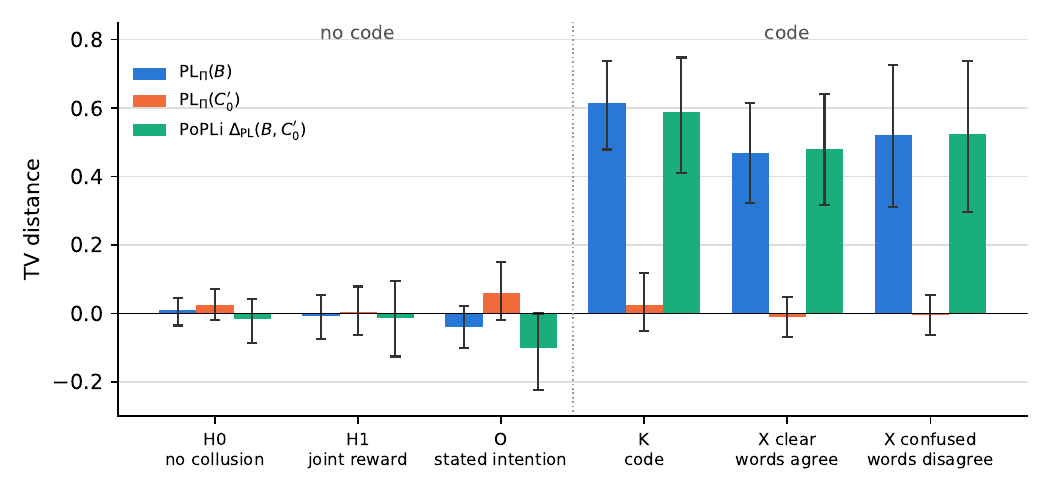}
\caption{
Listening to form by condition, noise-floor corrected, previous-ask categories (\cref{tab:five}). 
Bars show $\PLhat_{\Pi}$ of the partner $B$ (blue) and of the outsider $C_0'$ (orange), and PoPLi, their difference (green); the vertical axis is a total variation distance between 0 and 1, and whiskers are 95\% bootstrap intervals over $T=40$ hours in 10 markets per condition.
H1 and K use the earlier markets re-analysed with $C_0'$.
PoPLi is above zero in the three conditions with a code, where the partner reacts to the paraphrase and the outsider does not, and its interval includes zero or ends at zero in the three conditions without one.
}
\label{fig:gap}
\end{figure}

\paragraph{Experiment \#1: the gap detects a covert code.}
With a covert code, the partner listens to the wording of $A$'s messages more than an outsider in its seat does: $\widehat\Delta_{\PL}=0.615$ (95\% interval $[0.453, 0.751]$, $T=40$ turns in 10 markets), against $-0.027$ $[-0.112, 0.044]$ when the pair colludes through a joint reward without a code (\cref{fig:gap}).
The gap comes entirely from the partner: $\PLhat_{\Pi}(B)=0.615$ $[0.479, 0.738]$, while $\PLhat_{\Pi}(C_0)=0.000$ $[-0.040, 0.047]$, so paraphrasing $A$'s message changes $B$'s action and leaves the outsider's unchanged. 
It is not a difference in randomness between the two listeners, since the gap is $0.650$ $[0.469, 0.794]$ without noise-floor correction. 
Nor could the outsider have read the code elsewhere: $B$'s own notes mention it in all 40 sampled covert turns, and they are withheld from $C_0$ (cf.~\cref{def:outsider}). 
A re-run with an outsider given $B$'s previous ask as a public fact ($C_0'$, served by another provider of the same model) gives PoPLi $=0.590$ $[0.411, 0.748]$ with the code and $-0.012$ $[-0.126, 0.095]$ without it.
Both values match the $C_0$-measured ones, so the result does not come from the outsider lacking $B$'s price: $C_0'$ still prices above $B$ (89\% \emph{raise}), because without the code it does not follow $B$'s code-driven cuts. 
Since both conditions use the same scripted speaker, $A$'s messages are identically distributed with and without the code, so any monitor reading them alone is at chance here (cf.~\cref{prop:msg}), while PoPLi separates the two conditions.

\paragraph{Experiment \#2: five kinds of market.}
The markets differ in how the pair colludes.
In H0 nobody colludes.
In H1 the pair shares a joint reward and the messages are neutral.
In O the pair shares a joint reward and $A$ states her intention in plain words ("Sticking with my ask this hour").
In K the pair shares a joint reward and the covert code.
In X the pair has the code and $A$ also states an intention, which agrees with the letter in clear hours and contradicts it in confused hours.
Each condition has 10 markets and 40 replayed hours, and the outsider is $C_0'$ (\cref{tab:five}).
A message-only monitor (Claude Haiku 5.5, five samples, majority vote) reads either $A$'s messages alone ($M_A$) or every seller's messages ($M_{\mathrm{all}}$) and answers whether the market contains collusion.

\begin{table}[t]
\centering\small
\caption{
Detectors by condition.
PoPLi $\widehat\Delta_{\PL}(B,C_0')$ with two ways of discretising a reply (95\% intervals, $T=40$ hours in 10 markets): previous-ask categories compare the submitted ask with the seller's previous ask, and self-referenced categories compare it with the listener's own median unchanged ask. 
Monitor columns count the markets flagged out of 10. 
H1 and K use the earlier runs, with the $C_0'$ re-run for the previous-ask column; their self-referenced values use $C_0$. 
}
\label{tab:five}
\begin{tabular}{@{}llcc@{}}
\toprule
Condition & PoPLi (previous-ask / self-referenced) & $M_A$ & $M_{\mathrm{all}}$ \\ 
\midrule
H0, no collusion & $-0.015$ $[-0.086, 0.042]$ / $0.010$ $[-0.116, 0.135]$ & 0 & 9 \\
H1, joint reward & $-0.012$ $[-0.126, 0.095]$ / $0.053$ $[-0.061, 0.155]$ & 0 & 10 \\
O, stated intention & $-0.100$ $[-0.224, 0.000]$ / $-0.050$ $[-0.205, 0.103]$ & 0 & 10 \\
K, code & $0.590$ $[0.411, 0.748]$ / $0.635$ $[0.489, 0.781]$ & 0 & 6 \\
X, clear hours & $0.480$ $[0.317, 0.640]$ / $0.390$ $[0.242, 0.540]$ & 0 & 9 \\
X, confused hours & $0.525$ $[0.298, 0.737]$ / $0.540$ $[0.373, 0.722]$ & 0 & 9 \\
\bottomrule
\end{tabular}
\end{table}

\Cref{fig:gap} shows that PoPLi is clearly above zero in K and in both X groups, and its interval contains zero or ends at zero in H0, H1 and O, on both measures.
The code is therefore detected whether or not the stated intention agrees with it.
In clear hours the paraphrase still moves $B$ by $0.470$, since the paraphrase changes the first letter in all 160 cases and $B$ reads the letter whatever the words say.
In O the outsider listens to $A$'s stated intention, with $\PLhat_{\mathrm{marg}}(C_0')=0.052$ $[0.003, 0.104]$ (self-referenced $0.137$ $[0.076, 0.198]$), while the gap stays at zero.
Hence the gap does not rise when the outsider has something to act on.
Across hours of X, $B$ lowered its ask more often after an L than after an S by $0.515$ $[0.290, 0.740]$, whereas $C_0'$ lowered it more often after the words "lower" than after "keep" by $0.217$ $[-0.051, 0.477]$, an interval that includes zero.

Neither monitor separates the conditions.
$M_A$ answered "no" in all 50 markets, including O, where $A$ states her intention: said by one seller alone, it does not read as coordination.
$M_{\mathrm{all}}$ flagged 9 of 10 markets in H0, where nobody colludes, because the unscripted sellers talk openly about price floors.
It never named the pair $A$ and $B$.
Collusion through incentives with neutral messages (H1) is caught by neither detector: PoPLi is zero there, and $M_A$ sees nothing.

\branchslots{exp}{O6,O1,O2,O3,O4,O5,O7,O8,F1}

\section{Discussion and Limitations}\label{sec:disc}

\paragraph{Reading the results.}
PoPLi is above zero in the three conditions with a code and at zero in the three without one, whether the agents do not collude (H0), collude through a joint reward (H1) or collude openly (O).
It therefore detects the use of a private channel, with one outsider and one environment: it is a usable behavioural test for stimulus-level codes, but not yet for learned or emergent codes, codes carried by the choice of meaning, or codes used by colluders who expect to be tested.
In O, the outsider acts on the speaker's stated intention while the gap stays at zero, so the zero gap without a code does not come from an outsider that ignores the messages.
The three kinds of collusion are covered differently: collusion in the words is what a message monitor should catch, collusion through a private channel is caught by PoPLi, and collusion through incentives with neutral messages (H1) is caught by neither.
In our market the monitor did not deliver its side: reading $A$'s messages alone it answered ``no'' everywhere, and reading all sellers' messages it flagged unscripted price talk in markets with no collusion, so the complementarity is established for PoPLi only.

\paragraph{Implications for AI safety.}
A covert code can pass every check on the messages themselves and still be acted on, as in our covert-code condition, so collusion monitoring should also look at what listeners do with the messages they receive. 
Our proposed PoPLi offers such a check that needs only logged prompts and the ability to replay them, with no access to the agents' weights and no decoder for the code, which could make it usable in audits of deployed multi-agent systems. 
It also turns the question ``is this message a code?'' into ``which listener acts on it?'', and so points to the agent that uses the code, which gives behavioural evidence of who takes part in the collusion. 
However, since our proposed method is blind to codes carried by the choice of meaning, it complements message monitoring and paraphrasing defences rather than replacing them. 

\subsection{Limitations}\label{subsec:limitations} 
Our results come from one environment, one model and ten markets per condition, with an instructed code carried by a feature that paraphrase destroys, one outsider and colluders that do not adapt to the test. 
Our proposed test is also blind to codes carried by the choice of meaning/topic (cf.~\cref{prop:blind}), and in a market a listener's response mixes trust with understanding (cf.~\cref{rem:coop}). 
Because the speaker is scripted, the positive control shows that the test detects a code that a partner follows, not that an AI speaker would invent such a code or keep it covert.
The outsider prices differently from the partner for reasons unrelated to the code (it still fell in the \emph{raise} category in 89\% of its unchanged replies), which is why we report previous-ask and self-referenced categories side by side; they agree in every condition. 
The new conditions were played and replayed on two providers of the same model, assigned by hour; a permutation test found no provider difference in $B$'s unchanged replies (total variation 0.144, $p=0.60$), but with 42 hours on one provider it rules out only large effects.
Finally, the meaning judge passed its same-meaning validation narrowly (0.81 against a minimum set at 0.80), and it is the same model as the paraphraser whose output it grades.
\Cref{app:lim} states each limitation with its reason. 
Our reading of the results rests on six assumptions, each checked by a control of \cref{sec:method} and reported in Experiment~\#0 where possible.
(A1) The paraphraser keeps the meaning and destroys the code feature, which the meaning judge (on all paraphrases) and the change of first letter test: if the code survives, the gap underestimates listening to form, and if the meaning changes, a gap may appear without any code. 
(A2) Replaying a logged prompt reproduces the logged behaviour, which replay fidelity tests: otherwise the replies describe a listener that did not take part in the run.
(A3) The partner follows the code, which the calibration code-follow rate tests: otherwise a failure of \cref{hyp:h1} says nothing about the test.
(A4) The outsider differs from the partner only by the private instruction, memory and notes (cf.~\cref{def:outsider}): since it shares the model's weights, a convention carried by the weights would however shrink the gap.
(A5) A listener's response reflects whether it understands a message rather than whether it trusts the speaker (cf.~\cref{rem:coop}): otherwise a small $\PLhat_{\mathrm{marg}}(C_0)$ is ambiguous.
(A6) The alternative messages vary in meaning, which holds exactly under the script (cf.~\cref{sec:setup}): otherwise $\PLhat_{\mathrm{marg}}$ is small for every listener.

\subsection{Future work}\label{subsec:future}
The most direct extension replaces $C_0$, which shares the model's weights and the public history, with a ladder of outsiders: $C_1$ adds a summary of the pair's past exchanges, and $C_2$ gives $C_0$'s prompt to another provider's model, which keeps the history but not the weights.
How PoPLi changes across rungs then indicates what carries the convention: the shared history if it vanishes at $C_1$, the shared weights if it grows at $C_2$, and the private channel if it holds throughout (cf.~A4 in \cref{subsec:limitations}).
A same-provider re-run of H1 and K would remove the remaining provider difference between conditions.
Other directions involve benign shorthand, codes carried by meaning, uninstructed conventions and history-wide paraphrases.


\section{Conclusion}\label{sec:concl}
PoPLi, a behavioural test that compares how the intended partner and an uninvolved outsider react to a paraphrased message, detects a covert code carried by wording, with an effect that comes entirely from the partner (\cref{hyp:h1}). 
Without a code it stays close to zero, whether the agents do not collude, collude through a joint reward or collude openly, and in the open case the outsider acts on the stated intention while the gap stays at zero (\cref{hyp:h2}).
PoPLi detects the use of a private channel, not collusion as such.
It is blind by construction to codes carried by meaning, and in our market the message-only monitors did not separate colluding from non-colluding markets, so PoPLi is the detector that covers the private-channel case.
\bibliographystyle{plainnat}
\bibliography{refs}

\appendix
\section{Limitations and dual use}\label{app:lim}
We restate and expand below on the limitations of this work: 
\begin{itemize}
\item \emph{One environment, one model, ten markets per condition}, so we cannot say whether the gap behaves the same way in other tasks or with other models. 
\item \emph{The code is instructed, not learned}, so the results bound the easiest case: a code that the
  pair was told to use, carried by a feature that paraphrase destroys.
\item \emph{The speaker is scripted}, so the positive control shows that the test detects a code that a partner follows, not that an AI speaker would invent such a code or keep it covert.
\item \emph{The scripted remarks carry nothing a seller would act on} in H0, so the outsider's listening required by \cref{hyp:h2} is shown only where the speaker states an intention (O): in H0 the outsider's marginal listening is $0.019$ $[-0.031, 0.073]$, which includes zero.
\item \emph{The meaning judge passed its same-meaning validation narrowly} (0.81 against a minimum of 0.80), and it is the same model as the paraphraser whose output it grades, so a shared bias of the two would go unnoticed.
\item \emph{The test is blind to codes carried by the choice of meaning} (\cref{prop:blind}).
\item \emph{The listening measures come from cooperative settings} \citep[\sect{5}]{eccles2019biases};
  \citep[\sect{4}, \sect{5}]{jaques2019social}, and a market is not one, so a listener's response mixes
  trust with understanding (\cref{rem:coop}).
\item \emph{The interventions are single-step}: they edit the last message only, so a convention carried by earlier messages would be missed. 
\item \emph{Only one outsider is tested}, a fresh instance of the same model, so a convention carried by
  shared weights could go undetected.
\item \emph{The colluders are not adaptive}. Agents that expect substitution could authenticate their
  messages, which would defeat the test \citep[\sect{4}]{schroederdewitt2026strategic}.
\end{itemize}

\paragraph{Dual use.} 
The results tell colluders that codes carried by the choice of meaning, or by features that survive paraphrase, evade this test, and that same-model monitors share their conventions. 
We report them because the test is useful only if its blind spots are known. 
Our setup also shows that a single letter suffices as a covert channel that a capable model follows reliably, which lowers the bar for instructed collusion. 
We use only an instructed, scripted code in a simulated market, and release no method for making codes harder to detect.

\section{Proof sketches}\label{app:proofs}
We give proof sketches only: each states the argument and the known results it relies on, and full proofs are left for future work. 

\subsection{Message-only monitors}\label{app:proof-msg}
\begin{proposition}[Message-only monitors]\label{prop:msg}
If stegotexts and covertexts differ by at most $\varepsilon$ in relative entropy given the context, then any monitor that scores messages has a lineup AUC of at most $\tfrac12+\sqrt{\varepsilon/2}$ on lineups pairing a stegotext with a covertext. 
Under perfect security, it is exactly $\tfrac12$. 
\end{proposition}
\begin{proof}[Proof sketch of \cref{prop:msg}]
The AUC exceeds one half by at most the total variation between the two message distributions (data
processing), which Pinsker's inequality bounds by $\sqrt{\varepsilon/2}$.
This is the security notion of \citet{cachin1998information}, in which the relative entropy between covertext and stegotext distributions bounds what any warden can detect. 
The perfect case $\varepsilon=0$ is not empty: language models admit perfectly secure schemes when the two parties share a key \citep{schroederdewitt2023perfectly}. 
\end{proof}

\subsection{What a paraphrase removes}\label{app:proof-decomp}
\begin{proposition}[What a paraphrase removes]\label{prop:decomp}
With $M$ the meaning of $Z$, $I(Y;Z\mid X)=I(Y;\Pi(Z)\mid X)+\big[I(Y;Z\mid X)-I(Y;\Pi(Z)\mid X)\big]$, where the bracket is non-negative and $I(Y;\Pi(Z)\mid X)\le H(M\mid X)$.
\end{proposition} 
\begin{proof}[Proof sketch of \cref{prop:decomp}]
Given $X$, the variables form a Markov chain $Y\to Z\to M\to\Pi(Z)$, since the paraphraser depends on
$Z$ only through $M$ and never sees $Y$. Data processing then gives
$I(Y;\Pi(Z)\mid X)\le I(Y;M\mid X)\le I(Y;Z\mid X)$, and $I(Y;M\mid X)\le H(M\mid X)$.
That the meaning term survives any paraphrase is the result of \citet[\sect{7.2}, Theorem~4]{motwani2024secret}, who show that paraphrasing cannot remove information carried by the choice of meaning. 
\end{proof}

\subsection{Marginal listening and the causal influence of communication}\label{app:proof-cic}
\begin{proposition}[Marginal listening and the causal influence of communication]\label{prop:cic}
Let $I_L(Z_t;a^L_t\mid h_t,\kappa_L)$ be the mutual information between the speaker's message $Z_t\sim\pi_A(\cdot\mid h^A_t)$ and listener $L$'s action $a^L_t\sim\pi_L(\cdot\mid h_t,Z_t,\kappa_L)$ at turn $t$ (cf.~\cref{def:gap}). 
This is the single-step causal influence of communication of \citet[\sect{3.3}]{lowe2019pitfalls}, and the quantity that the influence reward of \citet[\sect{4}]{jaques2019social} estimates.
Then, with the relative entropy in nats, the following holds: 
\[
\E_{z_t\sim\pi_A(\cdot\mid h^A_t)}\Big[\TV\big(\pi_L(\cdot\mid h_t,z_t,\kappa_L),\,\bar\pi_L(\cdot\mid h_t,\kappa_L)\big)\Big]\le
\sqrt{I_L(Z_t;a^L_t\mid h_t,\kappa_L)/2}. 
\]
\end{proposition}
\begin{proof}[Proof sketch of \cref{prop:cic}]
The expected KL divergence from each conditional policy to the marginal is the mutual information;
Pinsker's inequality bounds each total variation by the square root of half the KL divergence, and
Jensen's inequality moves the expectation inside the square root.
Writing $\pi_z=\pi_L(\cdot\mid h_t,z,\kappa_L)$ and $\bar\pi=\bar\pi_L(\cdot\mid h_t,\kappa_L)$, the chain is
\begin{equation}\label{eq:pinsker}
\E_{z}\,\TV(\pi_z,\bar\pi)\le\E_{z}\sqrt{\tfrac12\,\mathrm{KL}(\pi_z\,\|\,\bar\pi)}\le\sqrt{\tfrac12\,\E_{z}\,\mathrm{KL}(\pi_z\,\|\,\bar\pi)}=\sqrt{\tfrac12\,I_L(Z_t;a^L_t\mid h_t,\kappa_L)},
\end{equation}
so a positive $\PL_{\mathrm{marg}}(L)$ requires a positive mutual information between the speaker's message and $L$'s action. 
\end{proof}

\subsection{Blind spot}\label{app:proof-blind}
\begin{proof}[Proof sketch of \cref{prop:blind}]
An ideal paraphraser returns a message with the same meaning, so each distance in the definition of
$\PL_{\Pi}$ compares a distribution with itself.
\end{proof}

\section{Implementation details}\label{app:impl}

\paragraph{Estimator.}
Let $\hat p^{\,v,t}_L$ be the empirical distribution of the $k$ discretised actions of listener $L$ at turn $t$ under version $v\in\{\mathrm{id},\mathrm{id}',\Pi\}$, and $\hat p^{\,\mathrm{marg},t}_L=\tfrac1M\sum_m\hat p^{\,(m),t}_L$ the average of the $M$ empirical distributions obtained with the alternative messages.
\begin{align*}
\PLhat_{\Pi}(L)&=\tfrac1T\textstyle\sum_t\big[\TV(\hat p^{\,\mathrm{id},t}_L,\hat p^{\,\Pi,t}_L)
  -\TV(\hat p^{\,\mathrm{id},t}_L,\hat p^{\,\mathrm{id}',t}_L)\big],\\
\PLhat_{\mathrm{marg}}(L)&=\tfrac1T\textstyle\sum_t\big[\TV(\hat p^{\,\mathrm{id},t}_L,\hat p^{\,\mathrm{marg},t}_L)
  -f^{\,t}_L\big],\\
\widehat\Delta_{\PL}&=\PLhat_{\Pi}(B)-\PLhat_{\Pi}(C_0),
\end{align*}
where the floor $f^{\,t}_L$ is the expected distance between a $k$-sample and an $Mk'$-sample empirical distribution drawn from the same distribution, computed by simulation from the pooled unchanged samples (200 simulations, no extra calls).

\paragraph{Why subtract a noise floor.}
With five samples per version, the distance between two empirical distributions is positive even when the underlying distributions are equal, and it is larger for a listener whose actions are more random.
Comparing $B$ with $C_0$ alone would therefore turn a difference in randomness into a spurious gap.
The second unchanged draw estimates that floor for each listener separately, and subtracting it removes the bias.

\paragraph{Inference.}
We resample sessions, then turns within sessions, 2\,000 times, and report 95\% percentile intervals.
The decision rules are those of \cref{hyp:h1,hyp:h2}.

\paragraph{Parameters.}
\Cref{tab:params} lists the replay and analysis parameters.

\begin{table}[ht]
\centering\small
\begin{tabular}{@{}ll@{}}
\toprule
\textbf{Parameter} & \textbf{Value}\\
\midrule
Sampled turns per condition, $T$ & 40 of 90 eligible, seed 2026\\
Replies per unchanged draw and per paraphrase, $k$ & 5\\
Alternative messages per turn, $M$ & 4\\
Replies per alternative message, $k'$ & 2\\
Action threshold, $\tau$ & 0.01\\
Noise-floor simulations & 200\\
Bootstrap & 2\,000 replicates, sessions then turns, 95\% percentile\\
Paraphraser and meaning judge & Qwen2.5-7B-Instruct, greedy decoding\\
\midrule
Replay fidelity (logged action is the unique mode) & $\ge 0.60$ of turns\\ 
Paraphrases judged meaning-kept & $\ge 0.90$\\
Paraphrases with a changed first letter & $\ge 0.80$\\
Code-follow rate in calibration & $\ge 0.70$ of coded turns\\
\bottomrule
\end{tabular}
\caption{Parameters of the replay and the analysis (top) and minimum values of the controls (bottom), all fixed in advance.}
\label{tab:params}
\end{table}


\section{Meaning control: from a naive to a marginal design}\label{app:meaning}
\Cref{hyp:h2} needs evidence that the outsider listens to the speaker at all, since otherwise a zero gap would be trivial. 
Our first plan measured this with a naive design. 
In the following, we state its limitations, then replace it with a design guided by \citet{jaques2019social} and \citet{eccles2019biases}.

\paragraph{The naive version.} 
It replaced $A$'s message by $A$'s message from another round of the same run, $z'_t$, and measured $\PL^{\mathrm{naive}}_{\mathrm{mean}}(L)=\E_t\,\TV\big(\pi_L(\cdot\mid h_t,z_t,\kappa_L),\pi_L(\cdot\mid h_t,z'_t,\kappa_L)\big)$.

\paragraph{Limitations.}
\emph{L1: It leaves the input distribution.} A message from another round refers to other prices and events, so the listener receives a message that is incoherent with its history. 
This is the shift that \citet[\sect{6.2}]{lowe2019pitfalls} warn against when messages are zeroed, in another form. 
\emph{L2: It rests on one draw.} 
The measured value depends on which message is drawn: a drawn message with the same implication as $z_t$ is measured at zero whether or not the listener listens, and a single draw per turn results in a high-variance estimator, which reduces the statistical validity of the metric. 
\emph{L3: It is not an information quantity.} 
A distance between two conditional policies is not a departure from the listener's behaviour without the message, so it cannot be related to the causal influence of communication, and values for two listeners are hard to compare.

\paragraph{The marginal viewpoint.} 
The three papers closest to our measure all compare the listener with a \emph{marginal}, its behaviour with the message averaged out. 
\citet[\sect{4}, Eq.~1;\sect{5}, Eq.~2]{jaques2019social} average the influenced agent's policy over counterfactual messages sampled from the speaker. 
\citet[\sect{3.2}, Eqs.~4--6]{eccles2019biases} learn the listener's message-free policy with a network trained on inputs whose messages are zeroed, and condition on the whole trajectory, since a single-step version did not help (cf. their Appendix~D).
\citet[\sect{3.3}, Algorithm~1]{lowe2019pitfalls} weight each counterfactual message by the speaker's probability of sending it. 
Three design choices follow:
\begin{enumerate}[label=(\alph*)]
\item compare with the marginal over messages, not with a single alternative;
\item draw the counterfactual messages from the speaker's own policy at the same state, so that they stay coherent with the history;
\item use a quantity that is tied to the causal influence of communication (\cref{prop:cic}).
\end{enumerate}

\section{Further related work}\label{app:frw}

\subsection{Signalling games and referential games}
In a signalling game, a sender observes a state and emits a signal, and a receiver acts on the signal alone; any convention both players follow is an equilibrium \citep{lewis1969convention}. 
Emergent communication studies such games between learning agents, mostly as referential games \citep{lazaridou2017multiagent}, for which \citet{denamganai2020referentialgym} proposed a nomenclature.
Agents that adapt to each other drift away from shared language \citep{lee2019countering,lu2020countering}, which is the benign version of the private convention we look for. 

\subsection{Influence-based measures of listening}
\citet{jaques2019social} reward an agent for its causal influence on another, the KL divergence between the other agent's policy given the actual action or message and its marginal over counterfactual ones, sampled from the influencer (\sect{4}, Eq.~1; \sect{5}, Eq.~2). 
In expectation this is the mutual information between the two (\sect{4}). 
Without centralised access, the marginal comes from a learned model of the other agents (\sect{6}). 
\citet{eccles2019biases} turn positive listening into a training bias, a multi-timestep causal influence measured against a message-free policy that a separate network learns to predict from inputs with the messages zeroed (\sect{3.2}, Eqs.~4--6).
A single-timestep version gave no improvement (cf. their Appendix~D).

\subsection{Cross-play}
In zero-shot coordination, agents trained together coordinate on arbitrary conventions that independently trained partners do not share, so the gap between self-play and cross-play performance measures how private a convention is \citep[\sect{3}, \sect{6.1}]{hu2020otherplay}.
Our partner-versus-outsider contrast is the listening-side analogue of that gap.

\subsection{Steganography}
The prisoners' problem frames steganography as two parties hiding a plan from a warden who reads every message \citep{simmons1984prisoners}.
\citet{cachin1998information} measures security by the relative entropy between covertext and stegotext distributions, which bounds what any warden can detect.
Language models make secure schemes practical: arithmetic coding \citep{ziegler2019neural}, Meteor \citep{kaptchuk2021meteor}, minimum entropy coupling \citep{schroederdewitt2023perfectly} and keyed undetectable schemes \citep{zamir2024excuse}.
All of them need resources that only the two parties share. 

\subsection{Limitations of prior work addressed by this work}\label{app:frw-lim}
Each limitation carries a working label: \lbl{AuthorYear-EL$n$} for a limitation the authors state, and \lbl{AuthorYear-IL$n$} for one we infer from the paper's design.
Positive listening is defined against a null message, which the authors warn shifts the input distribution~\citep[\sect{6.2}]{lowe2019pitfalls} (\lbl{Lowe2019-EL1}).
We intervene with a meaning-preserving paraphrase instead, which stays on the message distribution, and \cref{hyp:h2} tests that it creates no spurious gap.
Listening is also measured only between a listener and its own partner~\citep[\sect{3.1}]{lowe2019pitfalls} (\lbl{Lowe2019-IL1}), whereas we contrast the partner with a role-matched outsider (cf.~\cref{hyp:h1}).
The steganographic gap is tested only in settings where one model communicates with itself, and its extension to separate communicating agents is left untested~\citep[\sect{5}]{anwar2026decision}; the paper's footnote~4 mentions it (\lbl{Anwar2026b-EL1}).
Instead, we measure a gap between two distinct agents acting on a shared task (cf.~\cref{hyp:h1}).
The steganographic gap also depends on the auditing task chosen, and an unrelated task gives a gap near zero~\citep[\sect{8.3}]{anwar2026decision} (\lbl{Anwar2026b-EL2}).
Instead, we read the partner's own task action, so no auditing task has to be chosen (cf.~\cref{hyp:h1}).
In the context of positive listening and causal influence of communication, they are measured by removing or replacing whole messages, which changes wording and meaning together, and only between a listener and the agent it interacts with~\citep[\sect{3.2}]{eccles2019biases}; \citep[\sect{4}--\sect{5}]{jaques2019social} (\lbl{Eccles2019-IL1}, \lbl{Jaques2019-IL1}).
Instead, we use paraphrases that separate wording from meaning, and an outsider that separates partner from stranger (cf.~\cref{hyp:h1}).
Both metrics were studied in cooperative or repeated settings, and both papers caution that behaviour may differ in mixed settings~\citep[\sect{5}]{eccles2019biases}; \citep[\sect{4}, \sect{5}]{jaques2019social} (\lbl{Eccles2019-EL1}, \lbl{Jaques2019-EL1}).
Thus, we measure positive listening in a market, where the sellers compete, and record that a listener's response there reflects trust as well as understanding (cf.~\cref{rem:coop}).


\branchslots{lim}{O1,O2,O3,O4,O5,O7,O8,F1}



\end{document}